\documentclass[runningheads]{llncs}
\usepackage{eccv}

\usepackage{eccvabbrv}

\usepackage{graphicx}
\usepackage{booktabs}
\usepackage{multirow}
\usepackage[accsupp]{axessibility}
\usepackage{enumitem}
\usepackage[accsupp]{axessibility}  % Improves PDF readability for those with disabilities.

\usepackage{hyperref}
\usepackage{orcidlink}

\newcommand\mypara[1]{\vspace{1.0mm}\noindent\textbf{#1}}

\begin{document}

\title{An Empirical Study and Analysis of Text-to-Image Generation Using Large Language Model-Powered Textual Representation} 

\titlerunning{LLM-powered Text-to-Image Generation}

\author{
Zhiyu Tan\inst{1} \and
Mengping Yang \inst{2} \and 
Luozheng Qin \inst{3} \and
Hao Yang \inst{2} \and \\
Ye Qian \inst{2} \and
Qiang Zhou \inst{2} \and 
Cheng Zhang \inst{4} \and
Hao Li\thanks{Corresponding Author.} \inst{1} 
}

\authorrunning{Z. Tan et al.}

\institute{
{
\fontsize{8pt}{8pt}\selectfont
$^1$ Fudan University~
$^2$ INF Tech~
$^3$ Soochow University~
$^4$ Carnegie Mellon University\\
}
\smallskip\smallskip
\scriptsize{\texttt{8822tzy@gmail.com, \{yangmengping, yanghao, qianye.0514, zhouqiang\}@inftech.ai, 20225227060@stu.suda.edu.cn, czhang0528@gmail.com, lihao\_lh@fudan.edu.cn}} \\
}

\maketitle
\vspace{-20pt}
\begin{abstract}
One critical prerequisite for faithful text-to-image generation is the accurate understanding of text inputs. Existing methods leverage the text encoder of the CLIP model to represent input prompts. However, the pre-trained CLIP model can merely encode English with a maximum token length of 77. Moreover, the model capacity of the text encoder from CLIP is relatively limited compared to Large Language Models (LLMs), which offer multilingual input, accommodate longer context, and achieve superior text representation. In this paper, we investigate LLMs as the text encoder to improve the language understanding in text-to-image generation. Unfortunately, training text-to-image generative model with LLMs from scratch demands significant computational resources and data. To this end, we introduce a three-stage training pipeline, OmniDiffusion, that effectively and efficiently integrates the existing text-to-image model with LLMs. Specifically, we propose a lightweight adapter that enables fast training of the text-to-image model using the textual representations from LLMs. Extensive experiments demonstrate that our model supports not only multilingual but also longer input context with superior image generation quality. Project page: \url{https://llm-conditioned-diffusion.github.io}
\end{abstract}
\section{Introduction}
\label{sec:intro}

\begin{figure}[t]
    \centerline{\includegraphics[width=1\linewidth]{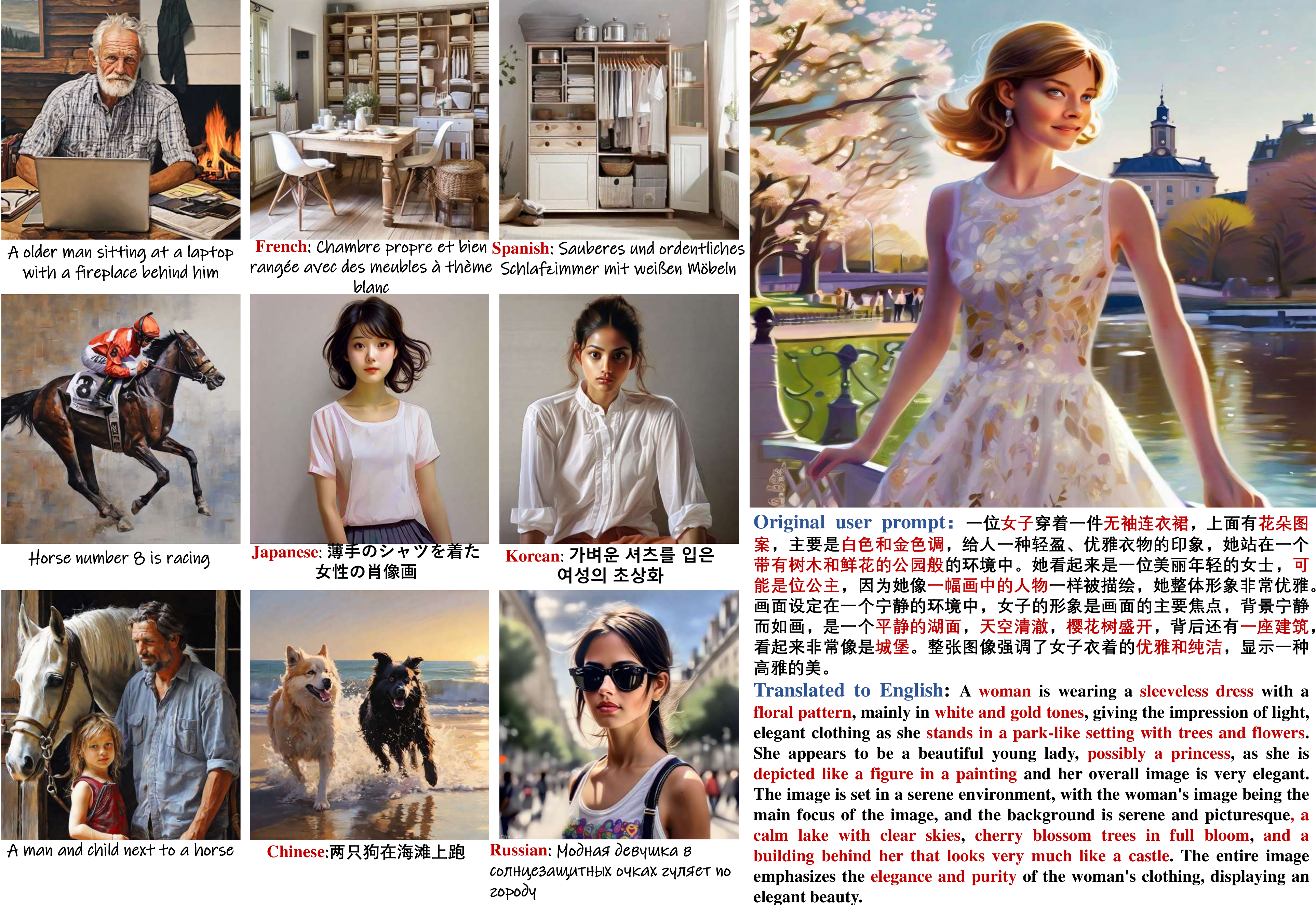}}
    \vspace{-2mm}
    \caption{\small Our proposed model could not only produce images with high visual quality given {English} input prompts ({left}), but also enables {multilingual} understanding capability for various language driven text-to-image generation ({middle}), as well as grasps {much longer contextual} information for generation ({right}).
    }
    \label{fig:teaser}
    \vspace{-2mm}
\end{figure}
 
 The recent advancement of proficient text-to-image generative models, such as DALL·E series~\cite{dalle1, dalle2, dalle3}, Imagen~\cite{imagen}, and Stable Diffusion~\cite{latentdiffusion}, have sparked a revolution in creating various images conditioned on texts.
Their technical breakthroughs not only push the boundaries of image synthesis but also significantly facilitate a spectrum of downstream applications, ranging from image editing and manipulation~\cite{sde, composer, controlnet}, personalized generation~\cite{ruiz2023dreambooth, arar2023domain, gal2022image},  and video generation~\cite{blattmann2023stable}.
In the context of text-to-image generation, a well-known fact is that the synthesis quality is greatly determined by the text features extracted from input prompts.
The root reason is that the generated images are bounded by the text representation capabilities of the text encoders.
Most existing text-to-image models~\cite{latentdiffusion, sdxl, dalle2} utilize CLIP~\cite{CLIP, openclip} to encode the input prompts.
Although CLIP nudges language understanding for text-to-image diffusion models, it has three essential drawbacks:
1) CLIP is tailored for English only, non-English native users have to translate their prompts before using image synthesis models, which may introduce extra inference latency and lose some contextual information of the prompts;
2) the max token length of CLIP is limited to 77, resulting information loss in longer text conditions;
and 3) the model capacity of CLIP is relatively small, leading to underperforming text representation capabilities, ultimately limiting the performance of text-to-image generative models.

In contrast, Large Language Models (LLMs)~\cite{achiam2023gpt, ouyang2022instruct-tuning, gpt3, gpt4} have shown unprecedented progress and gained extensive attentions in both academic and industry communities recently.
Attributed to ample training data, large-scale computation resources, LLMs herald a qualitative leap in text representation and language understanding abilities, which are remarkably better than those from CLIP models.
Moreover, since LLMs are trained on multilingual corpora with a longer preserved context length, it also support multilingual inputs and longer context accommodation. 
These advantages indicate that incorporating LLMs into text-to-image diffusion models could effectively address the limitations of CLIP models and significantly bootstrap the synthesis quality.

However, employing LLMs as the text encoders of diffusion models is challenging.
Unlike CLIP models that are trained to align features of image-text pairs, LLMs are developed solely from textual corpora, yielding text features that lack the awareness of visual information.
Therefore, directly employing LLMs as the text encoders of diffusion models might be inappropriate due to the information gap between textual and visual signals.
Additionally, training text-to-image generative model with LLMs from scratch is resource-consuming, requiring massive computational and data resources~\cite{imagen, dalle3, chen2023pixart}.
For instance, DALL-E 3~\cite{dalle3} is trained on $1.2$B images with a batchsize of $2048$ for $500,000$ iterations in total — a commitment of resources that many researchers and institutions cannot afford.

To address these issues, we propose a three-stage training pipeline, OmniDiffusion, that efficiently integrating LLMs into existing diffusion models.
The main idea of our method is a lightweight but effective adapter module to align the text features of LLMs with that of the visual-aware CLIP.
In this way, LLMs could capture the visual clues contained in the input prompts, thereby drive text-to-image diffusion models to produce appropriate images.
Specifically, we decompose the training procedure into three distinct stages. 
First, we adapt the features of LLMs into diffusion training process by aligning them with those from CLIP models, only adapter is optimized in this stage.
Then, we improve the synthesis quality through end-to-end text-image training.
After that, the aesthetic appeal of the generated images is enhanced by further finetuning on a carefully-curated dataset.
By doing so, the textual representation capabilities of LLMs can be fully activated and the model performance is well improved in terms of text alignment, synthesis quality and image aesthetics.
Notably, our model is trained with a fraction of the resources required by most text-to-image diffusion models while achieving superior synthesis quality and supporting multilingual input.

In order to verify the effectiveness of our proposed model, we conduct extensive empirical investigation on both English and Chinese prompts datasets, it turns out our model achieves favourable zero-shot FID/CLIP scores under various settings.
Besides, user studies demonstrate that our model could produce images that are preferred by human.
In summary, our main contributions are:

\begin{itemize}[itemsep=3pt,topsep=2pt,leftmargin=15pt]
\item We propose an effective approach for incorporating LLMs into text-to-image diffusion models, improving the awareness of LLMs towards the CLIP visual and textual space, thus facilitating more expressive language understanding.
\item We devise an efficient three-stage training pipeline, OmniDiffusion, that accomplish fast adaptation of LLM textual features with a small amount of resources, serving as an strong baseline of integrating LLMs into diffusion models and paving the way of this important topic.
\item Extensive experiments demonstrate that OmniDiffusion not only supports multilingual input conditions (\emph{i.e.,} Chinese, Japanese, Korean, \emph{etc.}) but also surpasses existing text-to-image diffusion models in terms of synthesis quality, text alignment and image aesthetics.
\end{itemize}

\section{Related Work}

\mypara{Text-to-image Diffusion Models.}
With the tremendous development of diffusion models~\cite{DDPM, ddim, nichol2021improved}, text-to-image generation~\cite{latentdiffusion, sdxl, dalle2, dalle3} have prompted appealing applications including image editing/manipulation~\cite{sde, composer, controlnet}, image translation/super-resolution~\cite{saharia2022palette, wu2023latent}, personalized generation~\cite{ruiz2023dreambooth, gal2022image, gal2022image}, \emph{etc}.
For instance, given a text condition, DALLE-2~\cite{dalle2} employed the joint embedding space of CLIP~\cite{CLIP} to learn a CLIP image embedding, based on which a decoder was trained to generate corresponding images.
GLIDE~\cite{glide} identified that leveraging classifier guidance~\cite{guideddiffusion, ho2022classifier} for text-to-image generation brings better synthesis quality.
Imagen~\cite{imagen} achieved better language understanding by leveraging a large transformer ({\emph{i.e.}, T5~\cite{T5}}) language models to encode text conditions.
Based on the text embedding power of CLIP~\cite{CLIP}, Latent Diffusion Model (LDM)~\cite{latentdiffusion}, also know as Stable Diffusion (SD)~\cite{stabilityai2023stable}, significantly improved the training efficiency of diffusion models by projecting samples into a pre-defined low-dimension latent space.
Further, DALL·E-3~\cite{dalle3} and SD-XL~\cite{sdxl} respectively improved text-to-image generation with better text captions, larger model capacity and novel conditioning schemes.
Notably, SD-XL opted both OpenCLIP~\cite{openclip} and CLIP to obtain more powerful text understanding ability.
Despite all these breakthroughs, most existing diffusion models are monolingual, \emph{i.e.,} can only understanding one specific language, thus hindering their further applications.
Moreover, the use of CLIP text encoder leads to very short contextual prompts understanding and underperforming expressive ability due to its maximum token length is only $77$ and the model size of CLIP is relatively small.

\mypara{Multilingual Text-to-image Generation.}
One straightforward way to produce multilingual images is to use machine translation tools to transform one language into another.
For instance, when using ERNIE-ViLG~\cite{zhang2021ernie} for text-to-image generation, one has to translate given prompts into Chinese before generation.
This solution, however, is often time-consuming and leads to unsatisfactory results due to the expression, grammar, and cultural differences.
To address this, several attempts have been made to enable multilingual diffusion models.
The CogView series~\cite{ding2021cogview, ding2022cogview2} extracted a bilingual vocabulary of Chinese and English tokens to support both of them for text-to-image generation.
Japanese SD~\cite{japanese_stable_diffusion} and Taiyi Diffusion~\cite{taiyi} respectively extended the English-only SD by incorporating corresponding language knowledge to facilitate Chinese and Japanese text conditioned synthesis.   
Taiyi-Diffusion-XL~\cite{wu2024taiyi} further improved the synthesis quality by integrating the capability of CLIP and SD-XL with bilingual continuous pre-training.
In order to enable more various languages, AltDiffusion~\cite{ye2023altdiffusion} first learned a multilingual text encoder and then re-train an initialized English-only diffusion model for concept alignment and quality improvement.

However, existing alternatives still have several limitations:
First, training bilingual/multilingual text encoders from scratch usually requires massive training data and cost.
Second, their language understanding abilities are still limited due to the misalignment between different languages and small text encoder model size.
Third, there is often a trade-off between language understanding and image quality because more attention is paid to textual alignment and the more the text-to-image diffusion model is ignored.
To address these issues, this paper proposes to leverage the text features of LLMs to provide promising language understanding ability and employ a resource-efficient three-stage training to ameliorate the synthesis quality, enabling a multilingual diffusion model without compromising the fidelity and diversity of generated images.

\mypara{LLMs for Downstream Tasks.}
The recent advancement of LLMs~\cite{ouyang2022instruct-tuning, gpt3, gpt4}  directly brings the blooming of visual content creation, multi-modal understanding, and various natural language processing tasks.
For instance, the emergence of GPT family, namely GPT-3~\cite{gpt3}, ChatGPT~\cite{chatgpt}, GPT-4~\cite{gpt4}, enabling convenient experiences for users to process tasks that were thought exclusive and burdensome before the era of LLMs.
Motivated by their great success, researchers seek to tailor the power of LLMs for various downstream tasks, such as interpreting visual signals including images~\cite{wu2023visual}, videos~\cite{li2023videochat}, \emph{etc.}
For example, LLaVA~\cite{llava, liu2023improved} and MiniGPT-4~\cite{zhu2023minigpt} utilized visual instruction learning to ameliorate the visual instruction following ability of LLMs.
Video-LLaMA~\cite{zhang2023video}, ChatVideo~\cite{li2023videochat} harnessed LLMs for video understanding tasks.
These attempts either design novel projectors to unify the representation spaces between different modalities or align the embedding space of pre-trained feature extractors, one typical example is LLaVA that joint a vision encoder (\emph{i.e.,} CLIP~\cite{CLIP}) and an LLM (\emph{i.e.,} Vicuna~\cite{chiang2023vicuna}) for multi-modal understanding.
However, in the context of text-to-image generation, the potential of LLMs remain less-explored.
Therefore, we present an effective practice of empowering text-to-image diffusion models with LLM-driven textual representation.
By aligning the text embedding space of LLMs and the visual/textual embedding space of CLIP via a lightweight adapter, we propose a multilingual text-to-image diffusion model that enables longer contextual understanding and high-quality generation.
\section{Method}

In order to fully explore the potential of LLMs for text-to-image generation, we propose a novel three-stage training pipeline.
First, a lightweight transformer-based adapter module is designed to align the text representations of LLMs with those of CLIP models, efficiently adapting the exceptional language understanding capabilities of LLMs for text-to-image generation.
Then, an end-to-end training scheme is conducted to further optimize the adapter and the pretrained UNet, improving the synthesis quality.
Finally, we perform a high-aesthetic finetuning on a small set of carefully curated high-quality images, improving the visual aesthetics of generated images.
In the following, we first present necessary preliminaries in~\cref{sec:preliminary}, followed by our overall framework in~\cref{sec:framework} and detailed description of our three-stage training pipeline in~\cref{sec:threestages}.

\subsection{Preliminaries}
\label{sec:preliminary}
\textbf{Diffusion Models} are trained to capture the distribution of training images through a sequential Markov chains of adding random noise into clean images and denoising pure noise to clean images.
Formally, the forward process is accomplished by adding random Gaussian noise ($\epsilon \sim \mathcal{N}(0, \mathbf{I})$) into clean images from training sets ($x_0 \sim {p}_{data}$):
\begin{equation}
x_t = \sqrt{\gamma(t)} x_0 + \sqrt{1-\gamma(t)} \epsilon,
\label{eq:add_noise}
\end{equation}
where $t \in [0,1]$ and $\gamma(t)$ is defined as a noise scheduler that monotonically descends from 1 to 0.
On the contrary, the reverse process aims to denoise Gaussian noises back to clean images by iteratively predicting the added noises at each reverse step:
\begin{equation}
L(\theta) = \mathbb{E}_{\epsilon \sim \mathcal{N}(0, \mathbf{I}), t}\left[ \left\| \epsilon - \epsilon_{\theta}(\mathbf{x}_t; t, c) \right\|^2 \right],
\label{eq:diffusion_loss}
\end{equation}
where $\epsilon_{\theta}$ denotes the denoising model parameterized by a neural network, and $c$ is the input conditions (\emph{e.g.,} class condition or text condition). 

\mypara{Text Condition Injection in Diffusion Models.}
The most common way to inject text conditions into diffusion models is interacting text representations and with image features via cross-attention mechanism~\cite{latentdiffusion, sdxl}.
Concretely, image features are utilized as the query (Q), and textual features are deployed as both the key (K) and the value (V) within the cross-attention layer.
The cross-attention layer computes a weighted sum of input sequences, where the weights are determined by the attention scores between Q and K:
$\text{Attention}(Q, K, V) = \text{softmax}\left({QK^T}/{\sqrt{d_k}}\right)V$.
When using CLIP text encoder to inject textual conditions, the input text initially undergoes tokenization, yielding a sequence of N tokens.
Note that if N exceeds a specified threshold T ($77$ for CLIP), the sequence will be truncated to T, and if N $<$ T, the sequence will be padded to T.
After that, the processed sequence is input into CLIP to extracts text feature, $h_\text{clip} \in \mathbb{R}^{T \times d}$, where $d$ signifies the dimensionality.
$h_\text{clip}$ is then used to interact with visual features via cross-attention mechanism, determining the content of generated images with respect to the text conditions.
It has been proved that the capacity of conditioned text feature plays a pivotal role in determining the quality of synthesized images~\cite{imagen, dalle3, chen2023pixart}.
Considering that LLMs~\cite{llama, baichuan} are trained on trillions of tokens with substantial model capacity for powerful language understanding, suggesting great potentials for text-to-image generation tasks.

\subsection{Framework}
\label{sec:framework}

\begin{figure}[t]
    \centering
    \includegraphics[width=.75\linewidth]{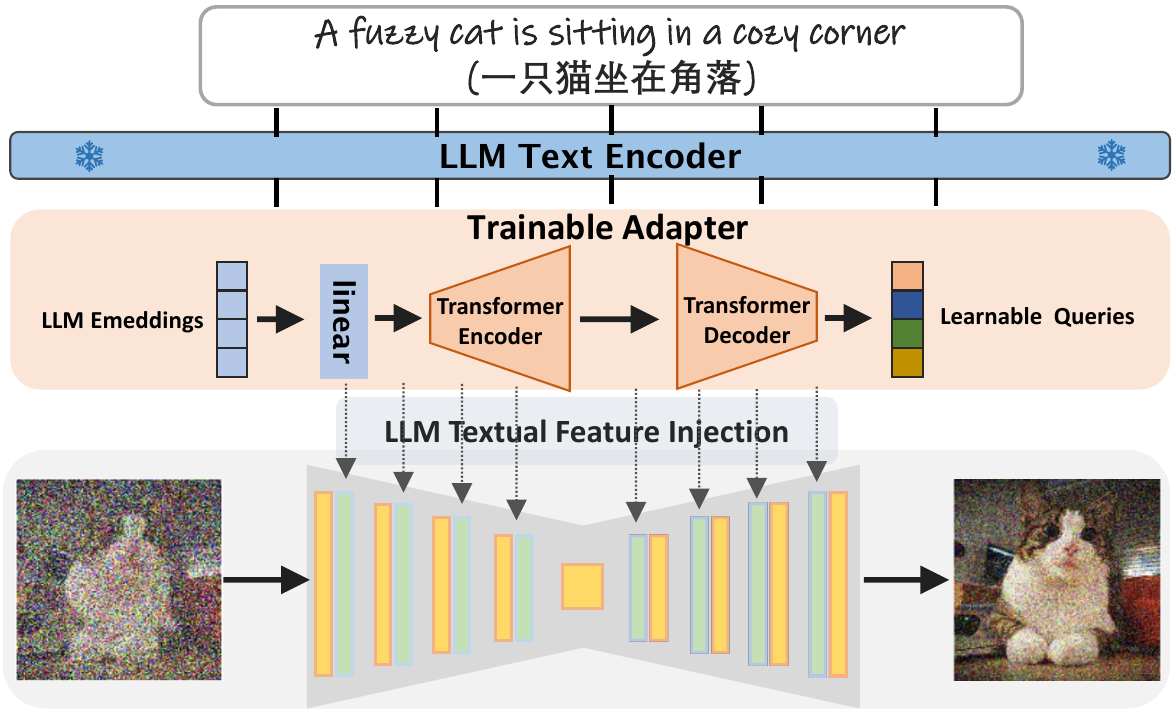}
    \caption{\small Overall framework of our proposed method. The lightweight adapter efficiently connects LLMs and diffusion models, enhancing diffusion models with more powerful language understanding ability.
    }
    \label{fig:framework}
\vspace{-8pt}
\end{figure}

Given the fact that the text features of LLMs are intrinsically ill-suited for text-to-image generation tasks, and directly training LLM-equipped diffusion models from scratch requires considerable resources, we derive an effective and efficient framework to adapt the textual representations of LLMs to diffusion models.
~\cref{fig:framework} presents the overall framework.
To be specific, we suffix an adapter module to the end of the LLM, achieving the alignment of text features between LLM and CLIP.
The adapter is a 4-layer encoder-decoder transformer with a learnable query sequence $q$, which is defined as:
\begin{equation}
h_\text{adapter} := \theta_\text{enc-dec}( \theta_\text{MLP}( h_\text{LLM} ), q ),
\label{eq:feature_mapping}
\end{equation}
where  $ h_\text{LLM} \in \mathbb{R}^{L \times d_\text{LLM} } $ and $h_\text{adapter} \in \mathbb{R}^{Q \times d_{q} }$, respectively denotes the output features of the LLM and the adapter.
$q$ is a hyper-parameter represents the number of learnable queries, which set as $227$ in implementation.

The output text features of the adapter-suffixed LLM $h_\text{adapter}$ serve as the text conditions of the diffusion training process, guiding the model to produce images that are consistent with input prompts.
In practice, we first project initial images into a low-dimensional latent space as $z_0$ with a pre-trained VAE~\cite{sdxl}, and obtain noisy latent features $z_t$ by adding noise $\epsilon$ to $z_0$.
Similar to the Latent Diffusion Model (LDM)~\cite{latentdiffusion}, we optimize our model by predicting the added noise conditioned on $h_\text{adapter}$ at given timesteps $t$:
\begin{equation}
L_\text{LDM}(\theta) = \mathbb{E}_{\epsilon \sim \mathcal{N}(0, \mathbf{I}), t}\left[ \left\| \epsilon - \epsilon_{\theta}(\mathbf{z}_t, t, h_\text{adapter}) \right\|^2 \right],
\label{eq:diffusion}
\end{equation}
where $\epsilon$ denotes the added noise and $\theta$ represents the model parameters.

\begin{figure}[t]
\centering
\includegraphics[width=1.0\linewidth]{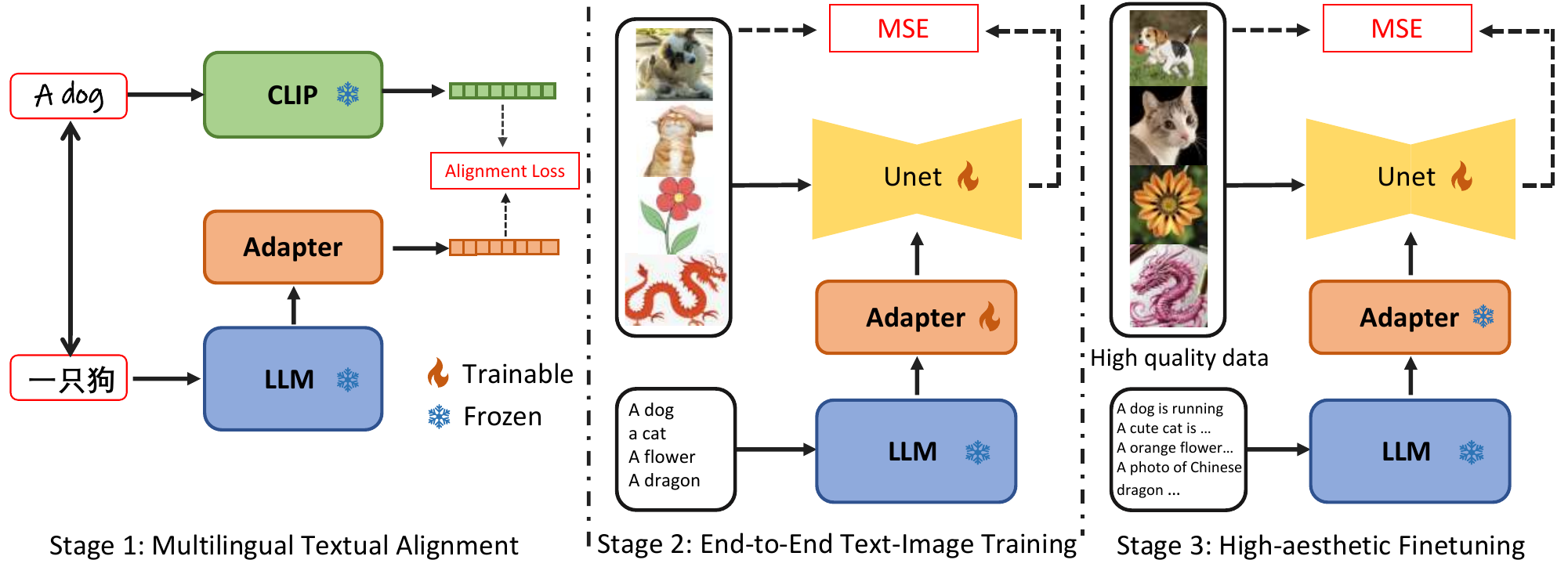}
\caption{\small Three-stage training pipeline. Multilingual textual alignment enables LLM to connect visual and textual information in the CLIP embedding space.
End-to-end text-image training explores the potential of LLM-derived textual features and improves generation quality.
High-aesthetic finetuning further ameliorates the visual aesthetic.
}
\label{fig2:training_pipeline}
% \vspace{-8pt}
\end{figure}

\subsection{Model Training Strategy}
\label{sec:threestages}

OmniDiffusion is trained with three stages, namely multilingual textual alignment, end-to-end text-image training, and high-aesthetic finetuning. The overall pipeline is presented in~\cref{fig2:training_pipeline}. We describe details as follows.

\mypara{Stage 1: Multilingual Textual Alignment.}
In order to introduce the text features of LLMs into the diffusion training process, we train the adapter to align the text feature of LLM with that of CLIP by merely optimizing the adapter through the proposed alignment loss.
During training, we use two types of text corpora, English-only and Chinese-English paired text.
For both of these datasets, the CLIP model is exclusively provided with English text, obtaining the features to be aligned, $h_\text{clip}$.
Meanwhile, the adapter-suffixed LLM is respectively fed with English and Chinese text for English-only and Chinese-English corpora, yielding the adaption-needed LLM-derived features, $h_\text{adapter}$.
Then, we align these features by optimizing the adapter module through an alignment loss, which is defined as:
\begin{equation}
    L(\theta_\text{adapter}) =  1 - \bigl\langle h_\text{clip}, h_\text{adapter} \bigl\rangle,
\label{eq:alignment_loss_cos}
\end{equation}
where $\bigl\langle \cdot, \cdot \bigl\rangle$ is the cosine similarity.
By minimizing the alignment loss, we can efficiently achieve multilingual textual alignment between LLM-derived features and CLIP image-text joint embedding space.
Moreover, we observe that the magnitudes of the token-level features extracted from CLIP and LLM differ significantly, which impairs the alignment performance(detailed experiment is provided in~\cref{sec:ablation}).
To address, we further incorporate a constraint on the token-level feature magnitudes of LLM and CLIP into the alignment loss:
\begin{equation}
    L(\theta_\text{adapter}) = \lambda * (|h_\text{clip}| - |h_\text{adapter}|)^2 - \bigl\langle h_\text{clip}, h_\text{adapter} \bigl\rangle,
\label{eq:alignment_loss_mc}
\end{equation}
where the first term constrains the magnitude of these two features should be aligned, and the second term ensures the features from the adapter-suffixed LLM align with the feature space of the CLIP model, $\lambda$ is the hyper-parameter that balances them.

Recall that the maximum token length of CLIP models is only $77$, resulting information loss on text that is longer than $77$.
Therefore, directly aligning the adapter-suffixed LLM with the CLIP may bias the LLM-derived text features to short prompt text-to-image generation and hinders alignment performance.
To mitigate this limitation, we adopt a segmented encoding approach.
It first segments the input prompt into chunks of $77$ tokens for separate CLIP encoding, the CLIP features of each chunk are then concatenated to obtain the complete feature.
Such that, the information beyond 77 tokens are well-preserved without modifying the CLIP model, allowing multilingual textual alignment on longer text data.

\mypara{Stage 2: End-to-end Text-image Training.}
Although stage 1 successfully aligns the textual feature space of the LLM to the CLIP embedding space, it does not fully explore the superior expressive capabilities of the LLM-derived text features, and the synthesis quality activated by LLMs could be further improved.
Therefore, to enhance image synthesis quality and fully exploit LLM features, this stage involves end-to-end text-image training on a curated internal dataset of $43$M text-image pairs.
During this stage, the parameters of LLM remain frozen, while the parameters of the adapter module and the UNet from the diffusion model are optimized by minimizing the diffusion loss, ~\cref{eq:diffusion_loss}.
Accordingly, the adapter is trained to further connect the textual representation space of LLMs and the text-to-image diffusion model, and the UNet is optimized to align the text conditions with the corresponding images, thus producing faithful and text-aligned images.

\mypara{Stage 3: High-aesthetic Fine-tuning.}
Stage 2 ensures the alignment of image-text embeddings of LLMs and diffusion process, but the overall image quality is relatively random and unstable.
In order to guide the model to exclusively produce images with highly visual aesthetics, we further finetune the diffusion model with a set of carefully curated high-quality images, which has shown appealing quality gains in the community~\cite{dai2023emu, chen2023pixart}.
Specifically, we filter $40$K exceptionally high-quality images from a large amount of filtered images, where the filtering process includes text and optical character recognition, aesthetic score filter, image-text alignment, position, and object detection, as well as human visual judgment.
Further, we employ a large multimodal model~\cite{llava} to caption these images for highly descriptive text conditions with specific styles.
In stage 3, the model is slightly optimized for nearly $5,000$ steps with a batchsize of $64$.

\begin{figure}
    \centerline{\includegraphics[width=1\linewidth]{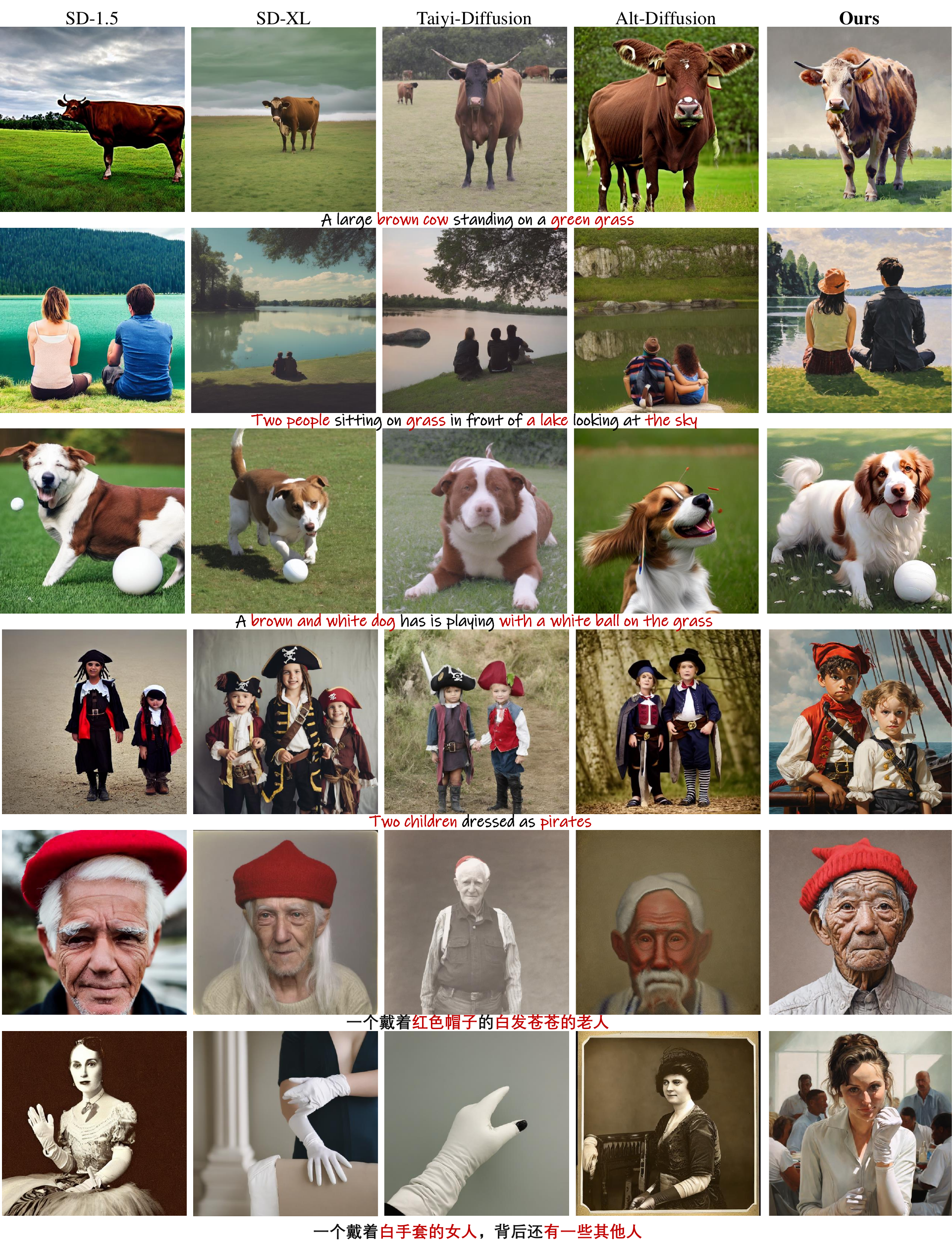}}
    \caption{\small Qualitative results of OmniDiffusion and competing methods.
    For models that do not support multilingual text conditions, we translate the given prompts into corresponding language to generate images.
    OmniDiffusion could produce images with accurate text-image alignment and higher visual quality.
    }
    \label{fig:sota_compare_showcase}
    \vspace{-3mm}
\end{figure}

\vspace{-5pt}
\section{Experiments}
We conduct qualitative and quantitative experiments to validate our method. We first introduce experimental setups in \cref{sec:implement}. We then compare the proposed OmniDiffusion with state-to-the-art models in \cref{sec:mainres}. Finally, we show ablative studies and analyses in \cref{sec:ablation}.

\subsection{Experimental Setup}
\label{sec:implement}

\begin{table}[t]
\centering
\small
\caption{\small Quantitative comparisons between our proposed method and existing baselines.
OmniDiffusion achieves the best or second-best results in FID and CLIP-Score, while outperforms compared methods significantly in aesthetics score (Aes).
}
\vspace{-2mm}
\resizebox{\textwidth}{!}{
\begin{tabular}{l|ccc|ccc|lll|lll} \toprule
\multirow{2}{*}{Model}         & \multicolumn{3}{c}{COCO-en} & \multicolumn{3}{|c|}{COCO-cn} & \multicolumn{3}{c|}{Flickr-30k-en}                                                 & \multicolumn{3}{c}{Flickr-8k-cn}                                          \\ %\cline{2-13} 
                               & FID     & CLIP-s  & Aes     & FID    & CLIP-s  & Aes   & FID   & CLIP-s  & Aes & FID     & CLIP-s  & Aes \\ \midrule
Taiyi-Diffusion                & \textbf{18.59}   & 0.2305  & 5.00    & \textbf{72.58}  & 0.3389  & 4.98  & 80.66 & 0.2200 & 4.99 & \textbf{78.00}  & 0.3412 & 4.98                 \\
Alt-Diffusion                  & 33.20   & 0.2427  & 5.49    & 89.41  & 0.3382  & 5.47  & 83.73 & \textbf{0.2522} & 5.47 & 88.38  & \textbf{0.3476} & 5.48                 \\ \midrule
\textbf{Ours}                  & 22.38   & \textbf{0.2463}  & \textbf{6.80}    & 74.65  & \textbf{0.3418}  & \textbf{6.62}  & \textbf{78.16} & 0.2504 & \textbf{7.03} & 83.53  & 0.3326 & \textbf{6.86}                 \\\bottomrule
\end{tabular}
}
\label{tab:compare_with_other_baselines}
% \vspace{-10pt}
\end{table}

\mypara{Datasets.}
Our three-stage training pipeline is respectively trained on pure text data (stage 1), text-image data (stage 2), and high-quality text-image data (stage 3).
Regarding the text data, we use the text captions of LAION-5B~\cite{schuhmann2022laion} and collected real-world user prompts.
For the paired text-image data for stage 2, we similarly employ the LAION-5B dataset and build an self-collected text-image dataset from internet, yielding a total of $43$M samples.
Notably, to fully exploit LLM-powered language understanding ability as done in~\cite{dalle3, chen2023pixart}, we re-caption these images with a self-trained multi-modal LLMs, providing more descriptive text conditions for training.
Finally, a carefully-crafted high-quality dataset is consist of both real-world images from internet and public-available image synthesis benchmarks~\cite{sun2024journeydb}, obtaining a total of $40$K images with a average aesthetic score of $7.51$.
%Notably, All of the aforementioned data is filtered, pre-processed and re-captioned with a carefully designed data cleaning process.

\mypara{Evaluation Metrics.}
Following existing philosophy of evaluating text-to-image generative models, we include three commonly-used automatic metrics, CLIP-score~\cite{hessel2021clipscore}, FID~\cite{FID} and aesthetics score~\cite{schuhmann2022laion} to evaluate the generated images in terms of text alignment, synthesis quality and aesthetic appeal.
Considering that OmniDiffusion features multilingual text conditions, we gather both English and Chinese prompts for a thorough investigation.
Specifically, for English-driven text-to-image generation, we use COCO~\cite{lin2014microsoft}, Flickr-30K~\cite{young2014image}, and Flickr-8K~\cite{hodosh2013framing}, and use COCO-cn~\cite{li2019coco}, Flickr-8K-cn~\cite{li2016adding} for Chinese driven text-to-image generation.
Note that we only utilize the text captions for producing images in inference, and Chinese-CLIP~\cite{yang2022chinese} is used for calculating CLIP-Score of Chinese benchmarks.
The original texts are used for computing CLIP-Score, and the original images are used as reference distribution for zero-shot FID evaluation.

To better investigate the image visual quality of OmniDiffusion, we further conduct an user study to identify the human preference of different models.
Presented pairs of images generated by different models, $10$ individuals are asked to select more visually appealing images.
To ensure the reliability of human evaluation, annotators are not aware of the image was produced by which model.

\mypara{Comparison Baselines.}
Quantitatively, we compare OmniDiffusion with several existing multilingual text-to-image diffusion models, namely Alt-Diffusion~\cite{ye2023altdiffusion}, and Taiyi-Diffusion~\cite{taiyi}.
Moreover, for those baselines that do not support multilingual input conditions, \emph{i.e.,} SD-v1.5~\cite{stabilityai2023stable} and SD-XL~\cite{sdxl}, we compare their qualitative results based on machine-translated text conditions, \emph{e.g.,} translating Flickr-8K-cn texts into English.

\mypara{Implementation Details.}
We use the open-source Baichuan2-7B~\cite{baichuan} as the integrated LLM, and build our 4-layer encoder-decoder adapter based on the SD-XL. All of the other details remain unchanged.
The features aligned in stage 1 are text features concatenated from OpenCLIP and BigCLIP, similar to that of SD-XL.
In stage 2, the UNet parameters are loaded from that of SD-XL and trained on image-text data with a batchsize of $3072$, and classifier-free guidance is used by randomly dropping 10\% input conditions.
The whole training process is performed on 128 NVIDIA A800 GPUs for approximately $2$ days.
%DeepSpeed is used for acceleration.

\subsection{Main Results}
\label{sec:mainres}

\mypara{Quantitative Results.}
~\cref{tab:compare_with_other_baselines} presents the comparison results between our proposed method and existing alternatives.
Owing to the aligned powerful language understanding of LLMs, OmniDiffusion achieves the best or the second-best FID/CLIP-Score performances under all tested benchmarks while gaining a significant improvement in visual aesthetics.
It indicates that our model is able to generate images with competitive synthesis quality, textual alignment and aesthetic appeal.

\begin{figure}[t]
    \centerline{\includegraphics[width=1.\linewidth]{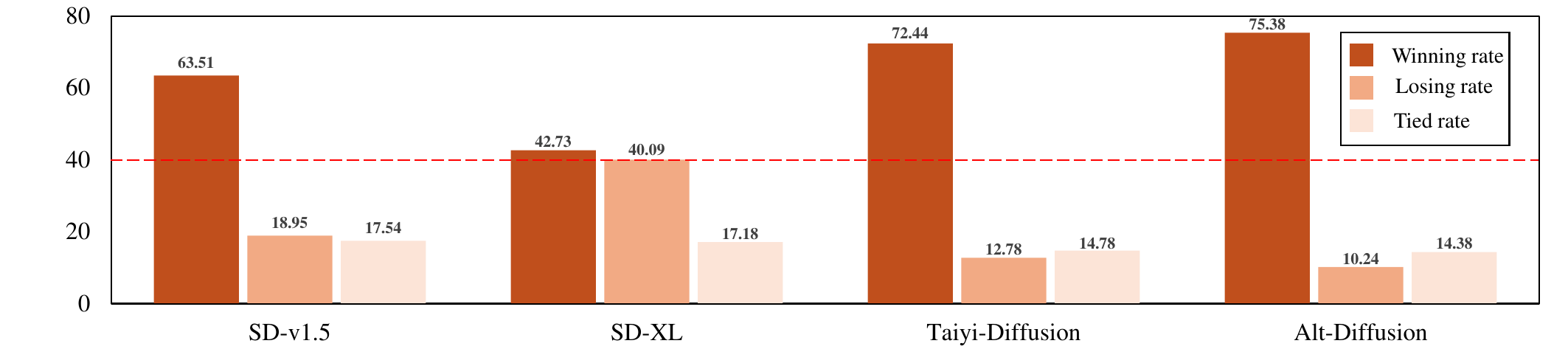}}
    \vspace{-3mm}
    \caption{\small Human evaluation results. Our model is consistently voted as the model that produce images with better visual quality.
    }
    \label{fig:humaneval}
    % \vspace{-8mm}
\end{figure}

\mypara{Qualitative Results.}
~\cref{fig:sota_compare_showcase} shows the visual outputs of our proposed method and comparison baselines, providing a clearer comparison on the synthesis quality.
Clearly, the overall quality of images generated by OmniDiffusion is more visually appealing, and the prompts following and text-image alignment are significantly better than others.
For instance, given ``\emph{A brown and white dog has is playing with a white ball on the grass}" as input, the image produced by our model is more realistic, the texture and visual details are more plausible.
Moreover, in the last row, compared models miss generating other people in the back of the woman, and their output images look unnatural.
This reflects that OmniDiffusion behaves better at following user prompts and synthesizing high-quality images.

\mypara{Human Evaluation.}
In order to investigate the human preference of OmniDiffusion, we conduct user studies to compare with prior approaches in a pair-by-pair manner.
Each time, we present users two images generated by OmniDiffusion and compared baselines, and then ask users to choose the better one.
~\cref{fig:humaneval} shows the results.
It turns out that our model is consistently identified as the better model when compared to different alternatives, particularly Taiyi-Diffusion and Alt-Diffusion.
Together with the above quantitative and qualitative results, the advance of our proposed method is well proved.

\subsection{Ablation Study and Further Analysis}
\label{sec:ablation}

\noindent \textbf{Ablating Different Training Stages.}
Here we remove different stages of our proposed three-stages training pipeline to testify the efficacy of each stage.
Specifically, we remove different training stages and remain other details unchanged, the quantitative results is shown in~\cref{tab:ablation_training_strategy}.
These results demonstrate that:
1) stage 2 effectively improve the FID scores, suggesting its efficacy of ameliorating synthesis quality;
2) stage 3 greatly contributes to the image visual aesthetics scores;
and 3) combining all stages could achieve the best trade-off in terms of visual quality, textual alignment, and aesthetics.
\begin{table}[t]
\centering
\small
\caption{\small Ablative results of different training stages.}
\vspace{-2mm}
\resizebox{0.9\textwidth}{!}{
\begin{tabular}{lccc|ccc|ccc|c} \toprule
\multirow{2}{*}{} & \multicolumn{3}{c|}{Training Stages} & \multicolumn{3}{c|}{COCO-en} & \multicolumn{3}{c|}{COCO-cn} & \multirow{2}{*}{GPU Days} \\ %\cline{2-10}
                               & stage 1 & stage 2 & stage 3 & FID    & CLIP-s & Aes  & FID    & CLIP-s & Aes  &       \\ \midrule
& \checkmark &            &            & 22.73  & 0.2423 & 5.77 & 115.72 & 0.2157 & 6.20 & 18       \\
&            & \checkmark &            & 15.50  & 0.2492 & 5.87 & 72.97  & 0.3311 & 5.49 & 53       \\
& \checkmark & \checkmark &            & 16.92  & 0.2662 & 5.95 & 71.58  & 0.3560 & 5.50 & 53       \\
& \checkmark &            & \checkmark & 44.11  & 0.2179 & 6.79 & 104.21 & 0.2731 & 6.78 & 18.2     \\
& \checkmark & \checkmark & \checkmark & 22.38  & 0.2463 & 6.80 & 74.65  & 0.3418 & 6.62 & 53.2     \\ \bottomrule
\end{tabular}
}
\label{tab:ablation_training_strategy}
% \vspace{-10pt}
\end{table}

\begin{figure}[t]
    \centerline{\includegraphics[width=\linewidth]{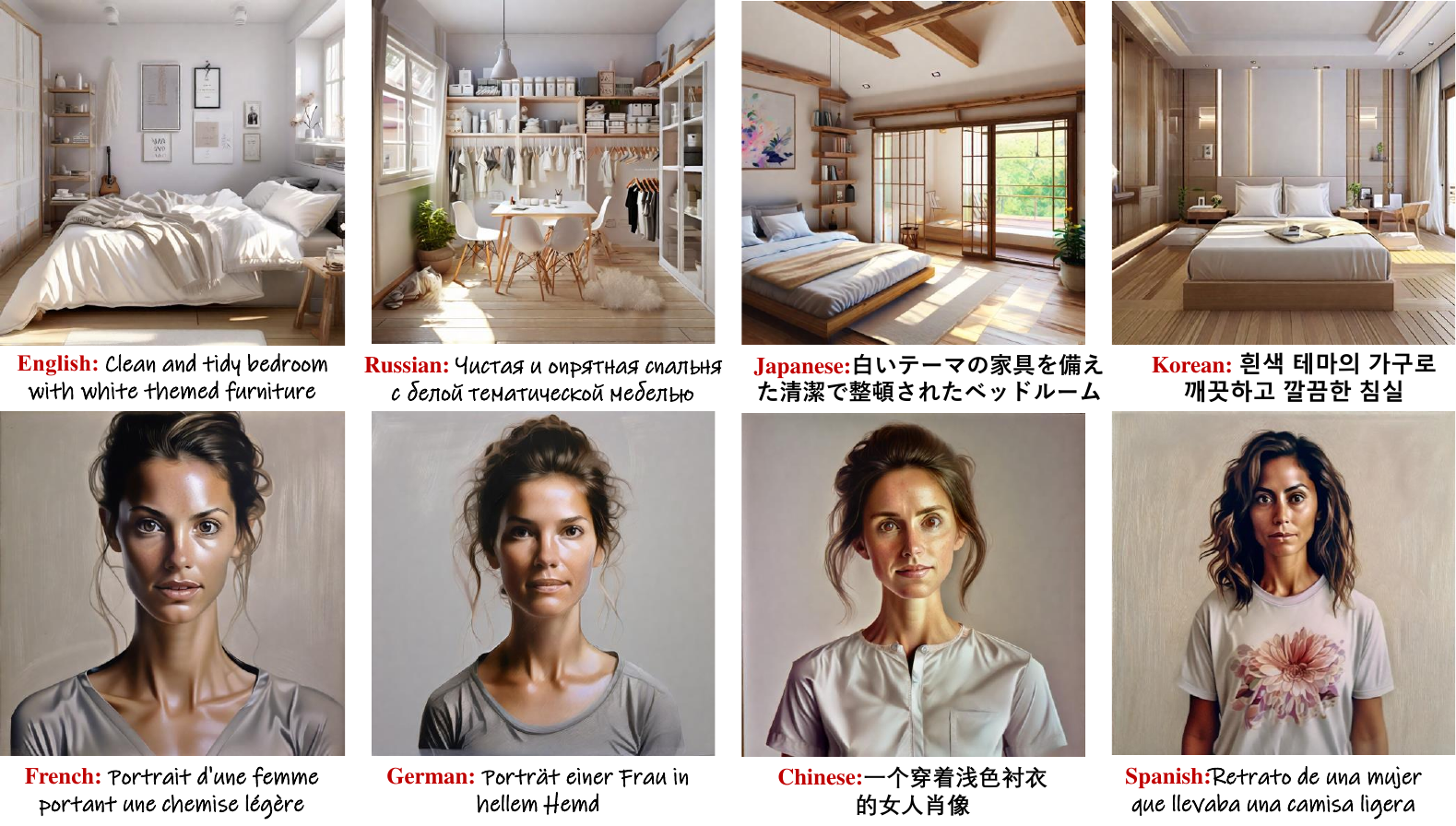}}
    \vspace{-3mm}
    \caption{\small Multilingual outputs.
    OmniDiffusion is capable of understanding various languages and produce corresponding image content.
    }
    \label{fig:multilingual}
    % \vspace{-12pt}
\end{figure}

\mypara{Ablating Magnitude Constraint.}
We added a magnitude to align the token of LLMs with CLIP text features, and here we provide the detailed analysis in ~\cref{tab:mag_diff}.
It could be observed that after applying the magnitude constraint, the token differences between text features of CLIP and that of LLMs are effectively mitigated, and the CLIP-Score is improved.
Such observations demonstrate the efficacy of explicitly aligning the encoded features of CLIP and LLMs.
\begin{table}[t]
\centering
\caption{\small The magnitude differences between LLMs and CLIP text features}.
\vspace{-4mm}
\begin{tabular}{c|cccccccc|c}
\toprule
Words       & A        & majestic & lion  & jumping & from & a & big & stone & CLIP-Score \\ \midrule
CLIP        & 12.88 & 18.5  & 17.25 & 20.88 & 20.76 & 23.75 & 17.63 & 21.00 & - \\
LLM w/ mc   & 14.44 & 18.23 & 18.14 & 21.87 & 21.86 & 20.76 & 21.03 & 20.84 & 0.2312 \\
LLM w/o mc  & 17.94 & 17.6  & 21.77	& 18.77	& 18.14 & 15.74	& 15.15	& 14.64 & 0.2242 \\ \bottomrule

\end{tabular}
\label{tab:mag_diff}
\vspace{4mm}
\end{table}

\mypara{Multilingual Text-to-image Generation.}
~\cref{fig:multilingual} provides the visual outputs of using various languages as the input conditions of our model.
Surprisingly, OmniDiffusion could understand these texts well and generate images with corresponding captions.
This amazing feature indicates that OmniDiffusion successfully integrates the powerful language understanding ability of LLMs into the text-to-image generation process, and fully exploit the potential of LLMs.

\begin{figure}[t]
    \centerline{\includegraphics[width=0.98\linewidth]{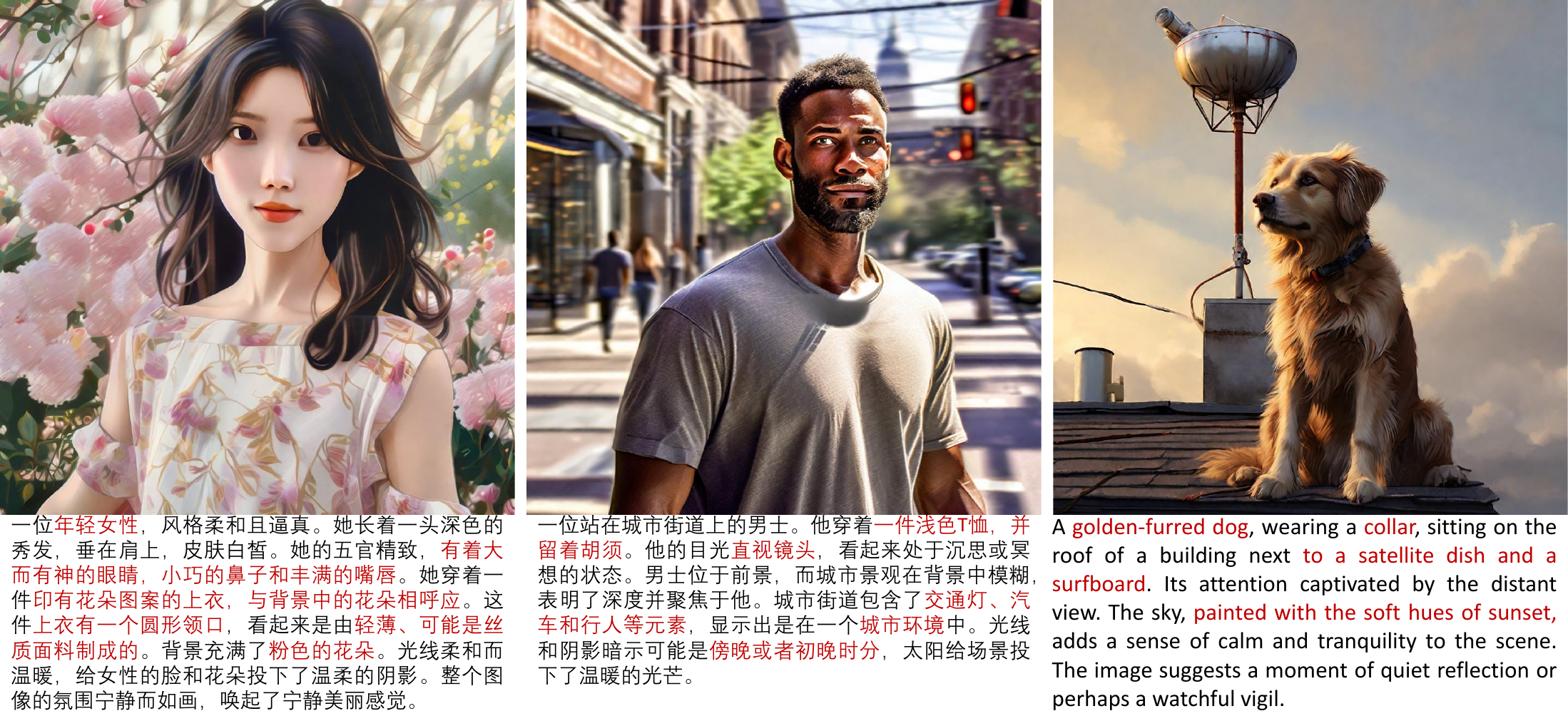}}
    % \vspace{-2mm}
    \caption{\small Long prompts driven generation results.
    OmniDiffusion can capture long contextual text semantics and synthesize text-alignment results.
    }
    \label{fig:longprompts}
    \vspace{-3mm}
\end{figure}

\mypara{Long Prompts Generation.}
To testify whether our model could understand long texts, we present the generated images under long prompts in~\cref{fig:longprompts}.
Remarkably, OmniDiffusion captures the meaning of prompts that are much longer than $77$ tokens and synthesizes images that well align with prompts, whereas prior methods usually fail under such setting.
This further reflects the powerful language understanding capability and synthesis quality of OmniDiffusion.

\section{Conclusion}
In this paper, we present an empirical study on exploiting the Large-language-model powered textual understanding capability for text-to-image diffusion models.
A lightweight but effective adapter is developed to efficiently connects LLMs' text features with CLIP's embeddings, making LLMs' representation applicable for T2I tasks.
Additionally, a novel three-stage training pipeline, OmniDiffusion, is designed to fully excavate LLMs' potential and simultaneous improve the synthesis quality.
Extensive results demonstrate the efficacy of each stage and the exceptional synthesis features enabled by OmniDiffusion, and our method provides an effective and practical baseline for connecting LLMs with T2I diffusion models.

\bibliographystyle{splncs04}
\bibliography{main}

\appendix
\setcounter{page}{1}
\newcommand{\AppendixPrefix}{A}
\renewcommand{\thefigure}{\AppendixPrefix\arabic{figure}}
\setcounter{figure}{0}
\renewcommand{\thetable}{\AppendixPrefix\arabic{table}} 
\setcounter{table}{0}
\renewcommand{\theequation}{\AppendixPrefix\arabic{equation}} 
\setcounter{equation}{0}

\section{Appendix}

This appendix is organized as follows:
First, we discuss the limitations and future works of the paper respectively in~\cref{supp-sec:limitations} and ~\cref{supp-sec:future}.
Next, we provide more analysis results to exhibit a more thorough perspective on the effectiveness of our approach in~\cref{supp-sec:analysis}.
Specifically, we compare the synthesis performance of our proposed three-stage training pipeline under different alignment losses, including Mean Square Error (MSE), cosine similarity, and cosine similarity with Magnitude constraint.
Additionally, trainable parameters are provided to demonstrate the efficiency of our proposed adapter.
Then, we exhibit details of the human evaluation in~\cref{{supp-sec:userstudy}}, including the user interface, user guidelines, and fine-grained results of user studies.
Finally, we provide more qualitative results of multilingual and long-prompt text-to-image synthesis, as well as comparison results with existing baselines in~\cref{supp-sec:quantitative}.
Moreover, we provide the qualitative visual results synthesized by different stages of our training pipeline, together with the quantitative results (See Tab.2) in the main paper, the effectiveness of our three-stage training pipeline is well identified.
Consistently, these qualitative results demonstrate that our proposed model effectively enables long-prompts following, multilingual, as well as better text-alignment understanding for text-to-image generation, providing users with satisfactory experience for producing visual content.

\section{Limitations}
\label{supp-sec:limitations}
Despite its effectiveness of powerful language understanding capability and efficiency of training pipeline, our proposed model still have several limitations:

\mypara{Human Evaluation.}
Conducting human evaluation can be time-consuming and resource-intensive, limiting the scale at which we can afford.
In our case, $10$ individuals with sufficient proficiency in both Chinese and English comprehension were involved to perform the human evaluation, each judged 400 image pairs produced by our proposed model and compared baselines.
Although we had tried our best to lower the difficulty of the human evaluation by organizing the evaluation task as a pair-by-pair comparison way, the results of which may be inevitably biased.
Additionally, human evaluation is inherently subjective and influenced by individual perspectives, biases, and preferences, particularly when it comes to the aspect of visual aesthetics.
As a consequence, evaluation on a different set of prompts, or with different users and guidelines may lead to different result, thus making our evaluation somewhat limited.

\mypara{Social Influence of Text-to-Image Models.}
Similar with other text-to-image models, our model can potentially generate images that contain offensive, inappropriate, or harmful content.
If the input text includes biased or discriminatory language, the model might generate images that perpetuate stereotypes or promote harmful ideologies.
Besides, our model may generate images that visually support false or misleading information provided in the prompt, potentially leading to the spread of misinformation or disinformation.
We involved a robust and unbiased dataset curation, created dedicated evaluation sets for bias detection and mitigation, and conducted adversarial testing through hours of redteaming, to alleviate these safety limitations.
Moreover, we believe that with appropriate use, our proposed model could provide users with interesting experiences for content recreation, and inspires more appealing research works to connect LLMs with diffusion models.

\mypara{Data-driven Finetuning.}
Following prior philosophy~\cite{dalle3, sdxl}, our model is trained on a large scale of image-text data to learn to fit the distribution of images conditioned on prompts.
However, bounded by the quality and diversity of the training data, some parts of the distribution are badly fitted.
Consequently, the model may struggle to generate certain objects that were not sufficiently learned during the data-driven finetuning.
Furthermore, our approach might not fully bring the powerful language understanding capabilities of LLM towards text-to-image synthesis.
Nevertheless, our approach strikes a considerable trade-off between efficiency and performance.

\section{Future Works}
\label{supp-sec:future}
Besides scaling up our models in terms of training data and learnable parameters, we plan to expand our works from three perspectives.
Firstly, we plan to try multiple adapter module and adopt the optimal one as the final module design of the adapter.
Secondly, we plan to find a more reasonable automatic metric to re-assess our model.
Thirdly, we plan to conduct quantitative evaluations on the multilingual text-to-image synthesis.

Although we have verified the effectiveness of our proposed adapter module, we simply utilized a 4-layer encoder-decoder transformer as the adapter module.
Consequently, We leave the exploration for the optimal adapter module design for our further research.

Besides, throughout our experiments, we found it difficult to find a suitable automatic metrics to assess our models and planned to solve it in the foreseeable future.
In our quantitative experiments, we utilized CLIP-score~\cite{hessel2021clipscore}, FID~\cite{FID}, aesthetics score~\cite{schuhmann2022laion} to automatically evaluate the synthesis quality.
Both the CLIP-score and FID do not correlate well with human assessment of the performance of generative models.
This might be caused by CLIP~\cite{CLIP} and Inception V3~\cite{szegedy2016inceptionv3}, models used to calculate CLIP-score and FID.
The CLIP and Inception V3 models are trained on low-resolution photos, which are significantly different from the image style preferred by our model.
Meanwhile, the image-text consistency between the captions and ground truth images of popular benchmarks, are not so highly correlated that it further flawed the metric. 
The lack of correlation of existing automatic metrics is also noticed by the image synthesis communities~\cite{sdxl, dai2023emu, kirstain2024pick}.

Due to the lack of a widely adopted multilingual benchmark for text-to-image synthesis, we were unable to perform quantitative evaluations on the multilingual text-to-image synthesis capability of our model.
In the future, we plan to deliver a multilingual text-to-image synthesis benchmark and quantitatively verify the multilingual image synthesis ability possessed by our model.

\section{More Analysis Results}
\label{supp-sec:analysis}

\mypara{Analysis on Alignment Loss Variants.}
During the textual alignment stage, we conducted an ablation study on three variants of alignment loss functions: Mean Square Error (MSE) Loss, cosine similarity loss, and cosine similarity loss with magnitude constraint.
Their formulations are illustrated below in~\cref{supp_eq:alignment_loss_mse}, \cref{supp_eq:alignment_loss_cos}, and~\cref{supp_eq:alignment_loss_cos_mc}, respectively.

\begin{equation}
    L(\theta_\text{adapter})_\text{MSE} =  (h_\text{clip}-h_\text{adapter})^2,
\label{supp_eq:alignment_loss_mse}
\end{equation}
\begin{equation}
    L(\theta_\text{adapter})_\text{cos} =  1 - \bigl\langle h_\text{clip}, h_\text{adapter} \bigl\rangle,
\label{supp_eq:alignment_loss_cos}
\end{equation}
\begin{equation}
    L(\theta_\text{adapter})_\text{cos}^* =  1 - \bigl\langle h_\text{clip}, h_\text{adapter} \bigl\rangle + (\text{mag}(h_\text{clip}) - \text{mag}(h_\text{adapter}))^2,
\label{supp_eq:alignment_loss_cos_mc}
\end{equation}
where $\bigl\langle  \cdot \bigl\rangle$ is the cosine similarity, $\text{mag}(\cdot)$ is the magnitude, $L(\theta_\text{adapter})_{cos^*}$ indicates the cosine similarity loss with magnitude constraint.

Specifically, we calculate CLIP-s on the validation set of the text alignment stage, the results of which are exhibited in Table \ref{tab:ablation_study_loss}.
As can be concluded, models trained with cosine similarity loss perform better in terms of CLIP text-image similarity across three types of corpora compared to those trained with MSE loss. 
The English CLIP-S score improved from 0.1688 to 0.2318, and the Chinese CLIP-S score increased from 0.1923 to 0.2086.
Furthermore, adding a magnitude constraint on top of the cosine similarity loss leads to even better alignment of the feature models. 

\begin{table}[h!]
\centering
\caption{\small Ablation study results for different loss functions. ``Baseline'' indicates the original SD-XL model that integrated with CLIP text encoders, ``$\text{Cosine Loss}^{*}$'' indicates our models optimized by cosine similarity loss with magnitude constraint.} 
\vspace{-4mm}

\begin{tabular}{l|ccc}
\toprule
\textbf{Model} & \textbf{CLIP-s} & \textbf{CLIP-s(en)} & \textbf{CLIP-s(ch)} \\
\midrule
Baseline                 & 0.2594  & 0.2868  & 0.2307 \\
MSE Loss                 & 0.1813  & 0.1688  & 0.1923 \\
Cosine Loss              & 0.2242  & 0.2318  & 0.2086 \\
$\text{Cosine Loss}^{*}$ & 0.2312  & 0.2482  & 0.2168 \\
\bottomrule
\end{tabular}
\vspace{-10pt}
\label{tab:ablation_study_loss}
\end{table}

\mypara{Analysis on Module Parameters.}
To quantitatively investigate the extra calculating budget brought by the adapter module, we summarize the parameter numbers of each module utilized in our model, and illustrated them in Table~\ref{tab:param_count}. 
Our model contains nearly 10.6B parameters, a substantial parameter number endows significant fitting ability on the distribution of high quality and aesthetic images conditioned on prompts.
The main contributor to the model parameters are the LLM text encoder~(7.B) and the UNet~(2.5B), together accounting for 94.6\% of the total number of parameters in the model.
Meanwhile, the extra introduced adapter is 487M, contributing merely 4.6\% of parameters in our model.
Therefore, the adapter module integrated in our model is quite lightweight, and the aforementioned quantitative experiments has confirmed its effectiveness.

\begin{table}[t]
\centering
\caption{\small Module Parameter Comparison. ``\#Parameters'' and ``Percentage'' respectively denote the number of module parameters and module parameter contribution percentage in model, respectively.
}
\vspace{-3mm}
\begin{tabular}{l|ccccc}
\toprule
\textbf{Module}    & \textbf{Our Adapter} & \textbf{VAE}    & \textbf{UNet}    & \textbf{LLM}     & \textbf{Total}    \\ \midrule
\#Parameters    & 487M    & 83M    & 2.5B    & 7.5B    & 10.6B    \\
Percentage (\%)   & 4.6   & 0.8  & 24.1  & 70.5  & 100.0 \\ \bottomrule
\end{tabular}
\vspace{-12pt}
\label{tab:param_count}
\end{table}

\mypara{Analysis on Text Encoders.}
To quantitatively verify the advantage of LLMs over other text encoders, we trained T5-integrated and LLM-integrated diffusion models with 14,000 steps of stage 2 training, we also include T5-based Wuerstchen for a fair comparison.
Considering that T5-integrated model understands English only while LLM-integrated model naively supports multilingual understanding.
Furthermore, T5-integrated model performs worse than LLM-integrated in FID and CLIP-s, indicating the advantage of LLMs.
\begin{table}[b]
\centering
\footnotesize
\vspace{-6pt}
\caption{Comparison with T5-empowered models. we trained T5-integrated and LLM-integrated diffusion models with the same training costs and procedure, we also include the popular T5-empowered diffusion model, Wuerstchen for a fair comparison.
}
\vspace{-3mm}
\begin{tabular}{l|ccc|ccc}
\hline
\multicolumn{1}{c}{\multirow{2}{*}{Models}} & \multicolumn{3}{|c}{Flickr8k-en} & \multicolumn{3}{|c}{COCO-en}\\
\multicolumn{1}{c|}{}     & FID   & CLIP-s & Aes  & FID   & CLIP-s & Aes  \\ \hline
\textbf{Ours (s2 T5)}     & 73.75 & 25.11  & 5.93 & 16.20 & 24.16  & 5.78 \\
Ours (s2 LLM)        & 72.72 & 26.23  & 5.97 & 15.50 & 24.92  & 5.87 \\
Wuerstchen (T5)           & 75.81 & 27.24  & 5.82 & 22.96 & 25.90  & 5.75 \\
\hline
\end{tabular}
\vspace{-10pt}
\label{supp-tab:text_encoders}
\end{table}

\mypara{Analysis on Training and Inference Costs.}
As for training costs, our model benefits from the proposed three-stages training paradigm that it only incorporates a lightweight adapter and is trained to efficiently connect LLMs with diffusion models.
Compared with the training complexity and resource requirements of training LLM-empowered diffusion models from scratch, our model achieves a better trade-off and is highly democratizing, serving as a strong baseline to deliver multilingual and long-prompts text understanding for T2I tasks.
As for the inference costs, we conduct experiments on diffusion models integrated with different text encoders and summarize the results in \cref{supp-tab:inference_costs}.
We can conclude that the main inference costs come from the inverse diffusion process, as the computational costs of integrating LLMs have minimal effect on the inference time.

\begin{table}[!ht]
\centering
\footnotesize
\tabcolsep 7pt % control the table width 
\vspace{-8pt}
\caption{Inference cost comparison with other text encoders. we include CLIP, T5, and a popular bilingual LLM, Baichuan2-7B, for comparison. 
In this table, we demonstrate the inference time under different text encoders and inference steps. 
As can be concluded, the inference time is highly consistent with the number of inference steps while showing little consistency with the employed text encoder. }
\vspace{-3mm}
\begin{tabular}{l|ccc}
\hline
\multicolumn{1}{c}{\multirow{2}{*}{Text Encoders}} & \multicolumn{3}{|c}{Inference Steps} \\
\multicolumn{1}{c|}{}   & 25    & 50    & 100   \\ \hline
CLIP                    & 3.13  & 6.04  & 11.84 \\
T5                      & 3.20  & 6.13  & 12.09 \\
Baichuan2-7B            & 3.21  & 6.16  & 12.03 \\ \hline
\end{tabular}
\vspace{-12pt}
\label{supp-tab:inference_costs}
\end{table}

\mypara{Analysis on Stage 1 Training.}
The average token length of the data used in stage 1 is 40.2, 11.7\% of which exceed 77.
During stage 1 training, we preserve the information beyond 77 tokens by applying a segmented encoding approach, thereby allowing multilingual textual alignment on longer text data.
In stage 1 training, the model also supports other languages, such as Japanese.
It is well-known in the community that, with well-pretrained multilingual embeddings, models align with one specific language also partially align with other languages.
Consequently, the model is trained on English and Chinese corpora in stage 1 training but preserved its multilingual understanding abilities.
However, to enhance model performance on other languages, a further stage 1 training on these language would be significantly beneficial.

\mypara{Analysis on Comparison with Other LLM-empowered Diffusion Models.}
There are some works that also explore how to enhance diffusion models using LLMs, such as ParaDiffusion~\cite{wu2023paradiffusion} and BiDiffusers~\cite{zhao2023bidiffuser}, which bear resemblance to us.
Here we clarify that this paper bootstraps text-to-image diffusion models with LLM-derived text features via suffixing a transformer-based adapter into LLMs.
Moreover, a novel and efficient three-stage training scheme to align LLMs with a pretrained diffusion model, enabling high-quality and high-aesthetic, as well as appealing prompt-following image synthesis, especially multilingual and long-prompts generation.
Notably, our work provides comprehensive experiments and analysis, namely automatic evaluation, user study, case study, and ablation study, serving as an valuable empirical study and strong baseline for LLM-integrated T2I diffusion models.

\begin{figure}[h]
\vspace{-3mm}
    \centerline{
    \includegraphics[width=0.95\linewidth]{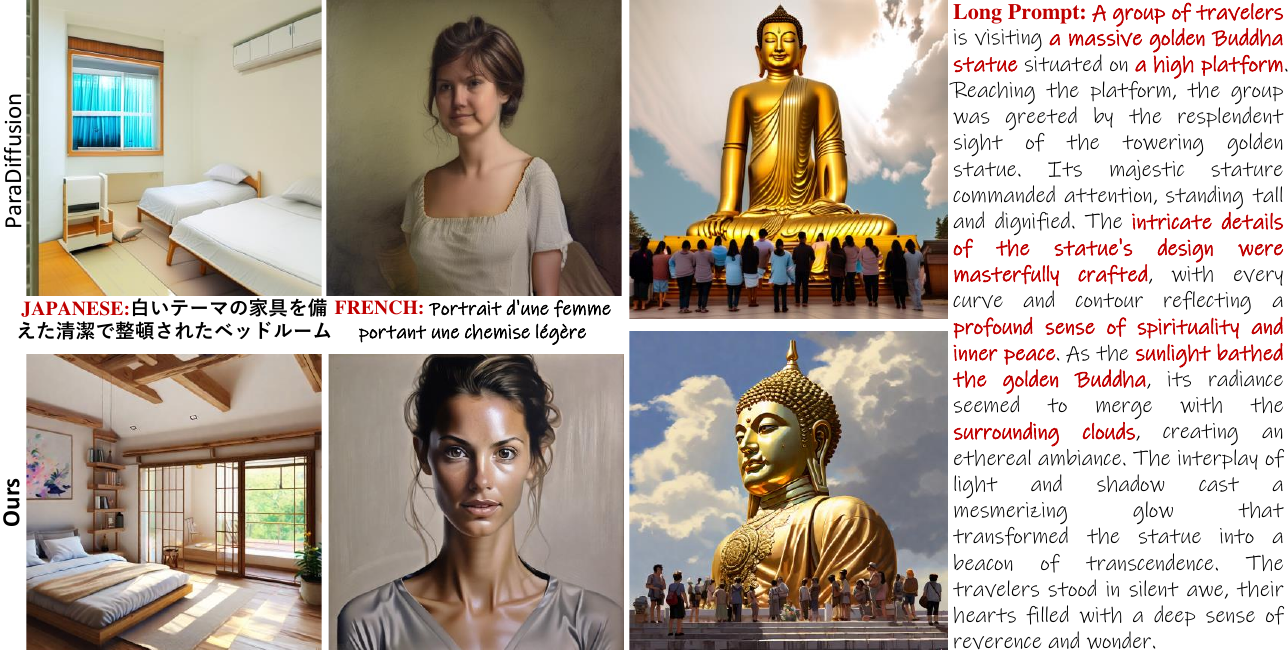}}
    \caption{\small Comparison with other LLM empowered diffusion models.
    BiDiffuser is not included because of the official released code is not executable.
    As can be seen, compared with ParaDiffusion, our model synthesizes images with higher quality, fidelity, aesthetic.
    }
    \label{fig:para_comparison}
    \vspace{-3mm}
\end{figure}

Further, our method differs Paradiffusion and BiDiffuser in the following aspects.
On the one hand, ParaDiffusion does not involve a textual alignment stage for faster convergence and is trained on almost 6.98 $\times$ of the data used to train our model.
Besides, ParaDiffusion use projection layer to connect a LoRA finetuned LLM with the diffusion models while we use a transformer adapter modules to connect a frozen LLM with the diffusion models.
On the other hand, BiDiffuser is constrained by its modified diffusion models and its training pipeline cannot be applied to other off-the-shelf diffusion pipelines such as SDXL and Pixart, while our methods do not require any modification to diffusion pipeline, which can be applied to any diffusion pipelines, even BiDiffuser.
Importantly, our empirical findings and practical solution~(See \cref{fig:para_comparison}) might inspire more interesting works to connect LLMs with text-to-image diffusion models in the community.

\section{Human Evaluation}
\label{supp-sec:userstudy}

\begin{figure}[h]
\vspace{-3mm}
    \centerline{\includegraphics[width=0.95\linewidth]{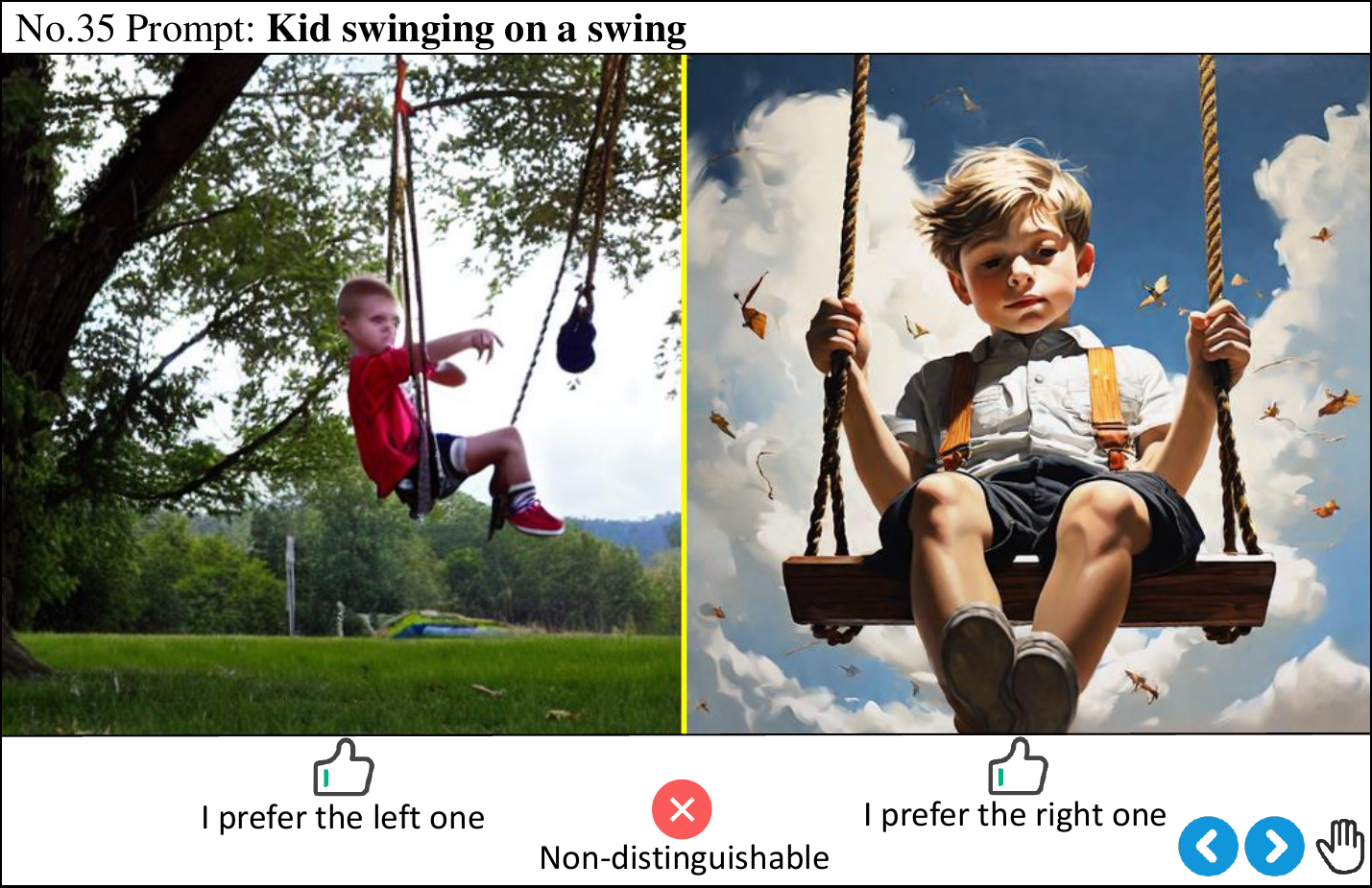}}
    \caption{\small Demonstration of our user interface.
    Each time, our specially designed user interface will provide a pair of images and the prompt used to generate them to the users.
    We incorporated six distinct icons to signify various functionalities of the user interface.
    }
    \label{fig:user_interface}
    \vspace{-3mm}
\end{figure}

The human evaluation was conducted in a pair-by-pair manner, with each user required to compare 400 identical image pairs.

\mypara{Data Preparation}
To prepare the data utilized for human evaluation, we first sample 400 prompts for image synthesis.
Specifically, each baseline model is allocated an equal distribution of 100 prompts, consisting of 50 Chinese prompts sampled from COCO-cn~\cite{li2019coco} and Flickr-8K-cn~\cite{li2016adding}, and 50 English prompts sampled from COCO~\cite{lin2014microsoft}, Flickr-30K~\cite{hodosh2013framing}, and Flickr-8K~\cite{hodosh2013framing}.
Then, the baseline models generate images guided by their respective prompts, while our models generate an image conditioned on the same prompt.
Subsequently, we create pairs between the images conditioned on the same prompt for comparison.
Overall, we prepared 400 pairs of images for human evaluation, with each pair composed of two images generated with the guidance of the same prompt, one by our model and the other one by baseline models.

\mypara{User Interface and Evaluation Procedure.}
To accomplish the evaluation effectively and efficiently, we designed an easy-to-understand user interface for users.
Figure \ref{fig:user_interface} demonstrates the functionalities of our user platform.
Each time, users are provided with a prompt and a pair of generated images.
Typically, users are required to click one of the two ``like'' buttons below each image, indicating their preference on a specific image.
If users are unable to distinguish a better image, they can click the red \textcolor{red}{X} button centrally below the images.
If users need to revise their previous compared pair, they can click the blue left and right arrow to jump to the previous and the next pairs.
In cases that users are confused about the evaluation, they can utilize the raise hand button to ask the help of the organizers.
Thanks to our adequate user training, reasonable evaluation procedure and friendly user interface, the raise hand button is set but not used in our evaluation.
Notably, the image pairs are disordered, thus users are blind to the models that generates the comparing images at the time.

\mypara{User Training and Guidelines}
Before performing human evaluation, we  made a series of efforts to guarantee it to be fair and unbiased.
First of all, we conduct a comprehensive user selection to select users that are willing to tolerate any inconvenient visual content that might be generated by the evaluated models.
Generally, 10 out of 33 users were selected to perform the evaluation, consisting of 3 graduate students major in text-to-image generation, three artist with sensitive to aesthetics, and four amateurs.
Then, to make users aware of the purpose and the standard for the human evaluation, we conduct a training over our carefully prepared user guidelines.
As exhibited in Table~\ref{tab:user_guidelines}, the user guidelines contain including instructions of user interface and detailed guidelines of human evaluation.
During the training, each user is provided with a copy of the user guidelines in Chinese, and is asked to understand these user guidelines.
We also explained to users the general purpose and standard for the human evaluation, and answer their questions about the user guidelines.
Notably, in cases that users might be confused about the compared pairs in progress, we add a raise hand button in the user interface so that users are able to ask for our help at any time.
Thanks to our adequate user training, reasonable evaluation procedure and friendly user interface, the raise hand button is set but not used.

\mypara{Fine-grained Results of Human Evaluation.}
We illustrate fine-grained results of human evaluation in Table~\ref{tab:fine_grained_human_evaluation}.
As can be seen, our human users reached agreement that our model obviously beats SD-v1.5, Taiyi-Diffusion and Alt-Diffusion.
As for SD-XL, the majority of the ten users agreed that the image generated by our model is slightly better than that by SD-XL.

\begin{table}[!t]
\centering
\caption{\small Fine-grained Results of User Studies. 
``Taiyi-Diff'' and ``Alt-Diff'' are abbreviation for ``Taiyi-Diffusion''~\cite{wu2024taiyi} and ``Alt-Diffusion''~\cite{ye2023altdiffusion}, respectively.
Our model gains much more votes by all participants when compared with various baselines, especially compared with multilingual models, \emph{i.e.,} Taiyi-Diffusion and Alt-Diffusion.
}
\vspace{-3mm}
\begin{tabular}{l|ccc|ccc|ccc|ccc}
\toprule
\multirow{2}{*}{\textbf{User ID}} & \multicolumn{3}{c|}{\emph{v.s} \textbf{SD-v1.5}} & \multicolumn{3}{c|}{\emph{v.s} \textbf{SD-XL}} & \multicolumn{3}{c|}{\emph{v.s} \textbf{Taiyi-Diff}} & \multicolumn{3}{c}{\emph{v.s} \textbf{Alt-Diff}} \\ \cline{2-13}
                         & Win     & Loss     & Tie    & Win    & Loss    & Tie    & Win        & Loss       & Tie       & Win       & Loss       & Tie      \\ \hline
No.1                     & 36      & 34       & 30     & 25     & 37      & 38     & 52         & 20         & 28        & 59        & 16         & 25       \\
No.2                     & 75      & 15       & 11     & 56     & 42      & 2      & 84         & 12         & 4         & 84        & 9          & 7        \\
No.3                     & 67      & 25       & 8      & 40     & 51      & 9      & 71         & 22         & 7         & 88        & 6          & 6        \\
No.4                     & 41      & 21       & 38     & 34     & 44      & 22     & 49         & 15         & 36        & 54        & 16         & 30       \\
No.5                     & 66      & 9        & 25     & 54     & 24      & 22     & 76         & 6          & 18        & 79        & 3          & 18       \\
No.6                     & 77      & 17       & 6      & 42     & 46      & 12     & 80         & 12         & 8         & 77        & 15         & 8        \\
No.7                     & 72      & 13       & 16     & 54     & 26      & 20     & 86         & 3          & 11        & 87        & 2          & 11       \\
No.8                     & 66      & 29       & 5      & 32     & 52      & 16     & 77         & 12         & 11        & 86        & 12         & 2        \\
No.9                     & 69      & 17       & 14     & 53     & 33      & 14     & 77         & 13         & 10        & 78        & 9          & 13       \\
No.10                    & 64      & 20       & 16     & 43     & 39      & 18     & 78         & 11         & 11        & 83        & 10         & 7        \\ \bottomrule
\end{tabular}
\vspace{-12pt}
\label{tab:fine_grained_human_evaluation}
\end{table}

\begin{table}[t]
\footnotesize
\centering
\caption{\small
User Guidelines of the Human Evaluation.
Considering that our users are Chinese speakers while our readers may not be, each user is provided with a copy of Chinese version of the user guidelines. 
Meanwhile, we demonstrate its translated English version in the following.
}

\begin{tabular}{|p{1\columnwidth}|}
\bottomrule
\textbf{User Guidelines}
\\ \hline
\textbf{Part I Guidelines of the User Interface} 

1.You are supposed to compare 400 pairs of images on the provided user interface throughout the whole evaluation process.

2.Each time, the user interface will show you a sequential number, a pair of images generated by different models, a prompt used to guide the generation of the images, and six different buttons to signify various functionalities of the user interface.

2.The sequential number for the comparing pairs printed at the top left corner of the user interface, is used to remind you of your working progress.

3.To accomplish the comparison of presenting image pairs, you are supposed to click on the ``like'' icon below a specific image. It indicates your preference for that specific image. 

4.In cases where you likes or dislikes both images, you can click the red ``X'' button located centrally below the images.

5.Once you perform guideline 3 or guideline 4, the user interface will automatically switch to the next pair.

6.In the bottom right corner of the interface, there are a left and a right arrow icons, which enable you to navigate to the previous and next image pairs and revise your comparison results.

7.If you are unsure of anything about the human evaluation, feel free to click the raise hand button in the bottom right corner of the user interface, we will be glad to help you.
\\ \hline
\textbf{Part II Guidelines of the Human Evaluation}

1.In general, you are supposed to compare the image pairs with the criterion of your personal preference.

2.If you are unsure about how to form your personal preference on the compared images. We suggest three criterion, image quality, text-image consistency and aesthetics.

3.Image Quality: whether the generated image is clear, details of the generated image objects are consistent with that in real world.

4.Text-image Consistency: whether the generated image is consistent with the description of the given prompt.

5.Aesthetics refers to the overall aesthetic quality of a generated image. It combines various visual elements such as color, shape, texture, and composition that creates a fascinating image as a whole. 

6.You are required to first compare 30 pairs of images to form a stable and reasonable assessment standard. Then, compare the whole 400 pairs of images from start to finish.

7.If you feel confused at anything about the human evaluation, feel free to click the raise hand button in the bottom right corner of the user interface, we will be glad to help you.

8. If you finished the comparison of all 400 pairs, you can submit your evaluation results by clicking the right arrow button.

9. Once you have submitted your evaluation results, we are very thankful to inform you that you have finished your job. Thank you once again for your contribution to our project.
\\ \hline
\end{tabular}
\label{tab:user_guidelines}
\end{table}

\clearpage
\section{More Quantitative Results}
\label{supp-sec:quantitative}
\begin{figure}[h!]
    \centerline{\includegraphics[width=0.95\linewidth]{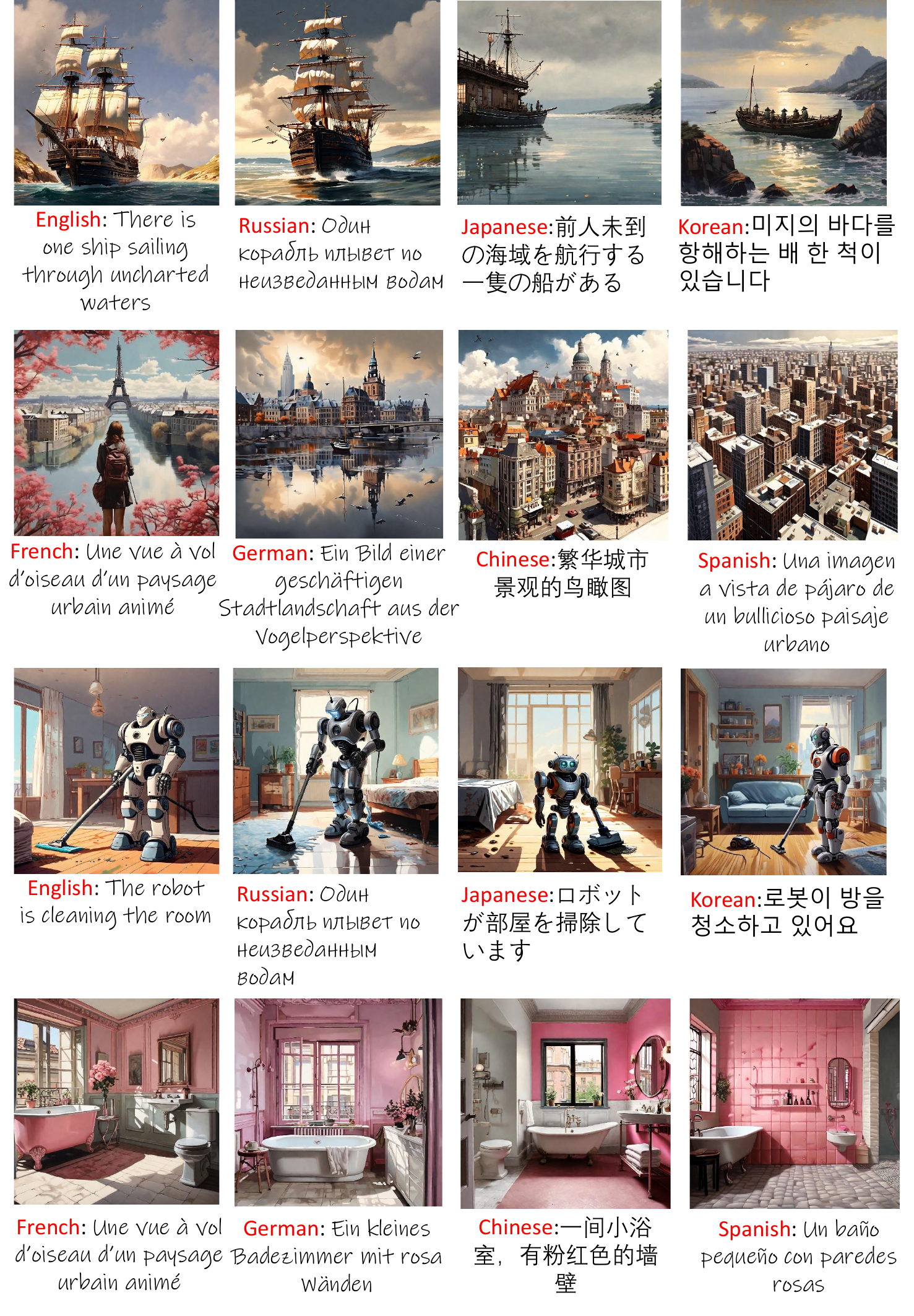}}
    \vspace{-4mm}
    \caption{\small Multilingual outputs.
     More examples showcasing that our model is capable of understanding various languages and produce corresponding image content.
    }
    \label{fig:multilingual_sp}
\end{figure}

\begin{figure}[h]
    \centerline{\includegraphics[width=1.0\linewidth]{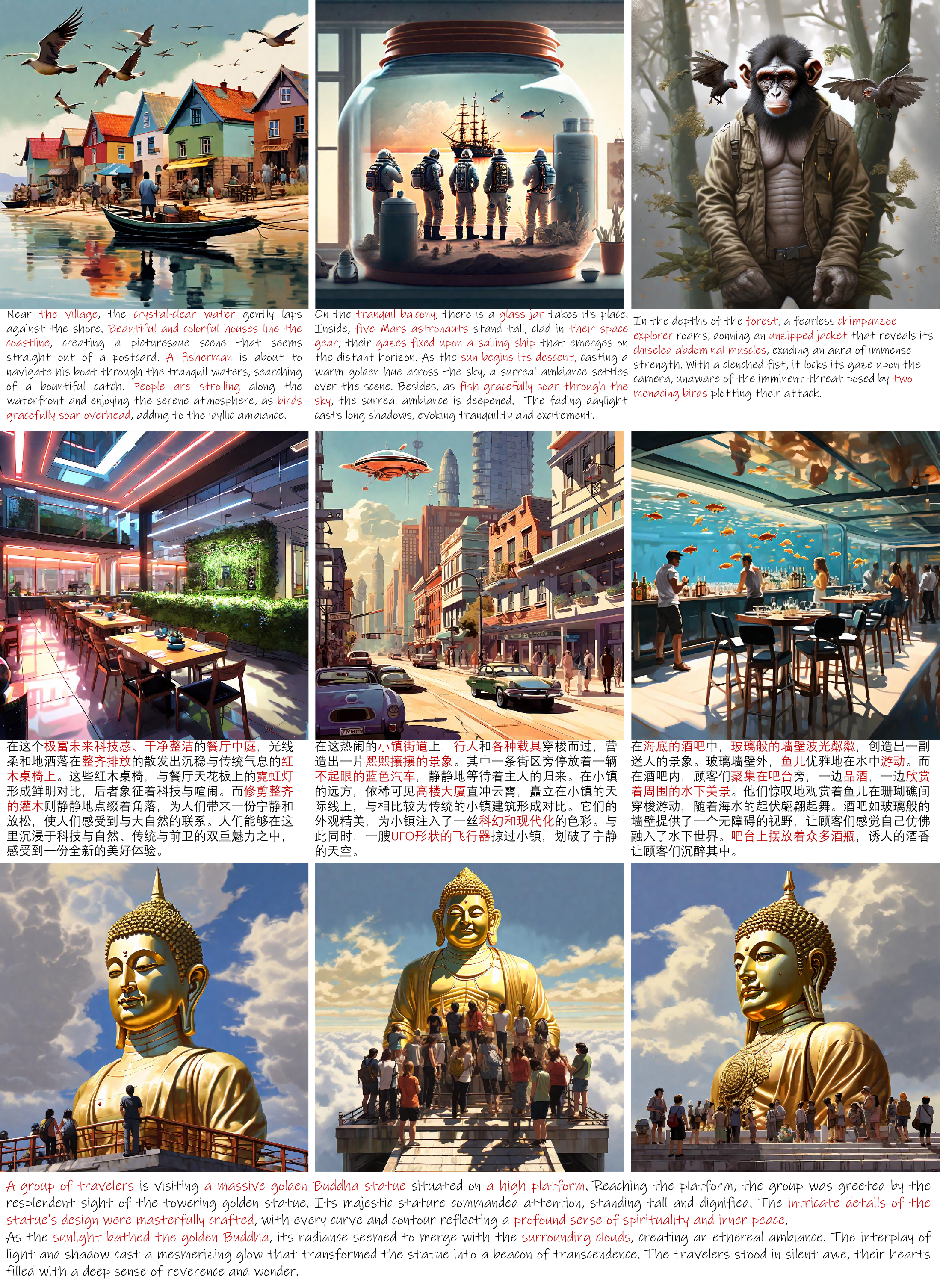}}
    \vspace{-3mm}
    \caption{\small Long prompts driven generation results. More examples showcasing that our model could capture long contextual text semantics and synthesize text-alignment results.
    Zoom in for details.
    }
    \label{fig:longprompts_sp_1}
    \vspace{-12pt}
\end{figure}
\begin{figure}[h]
    \centerline{\includegraphics[width=1.0\linewidth]{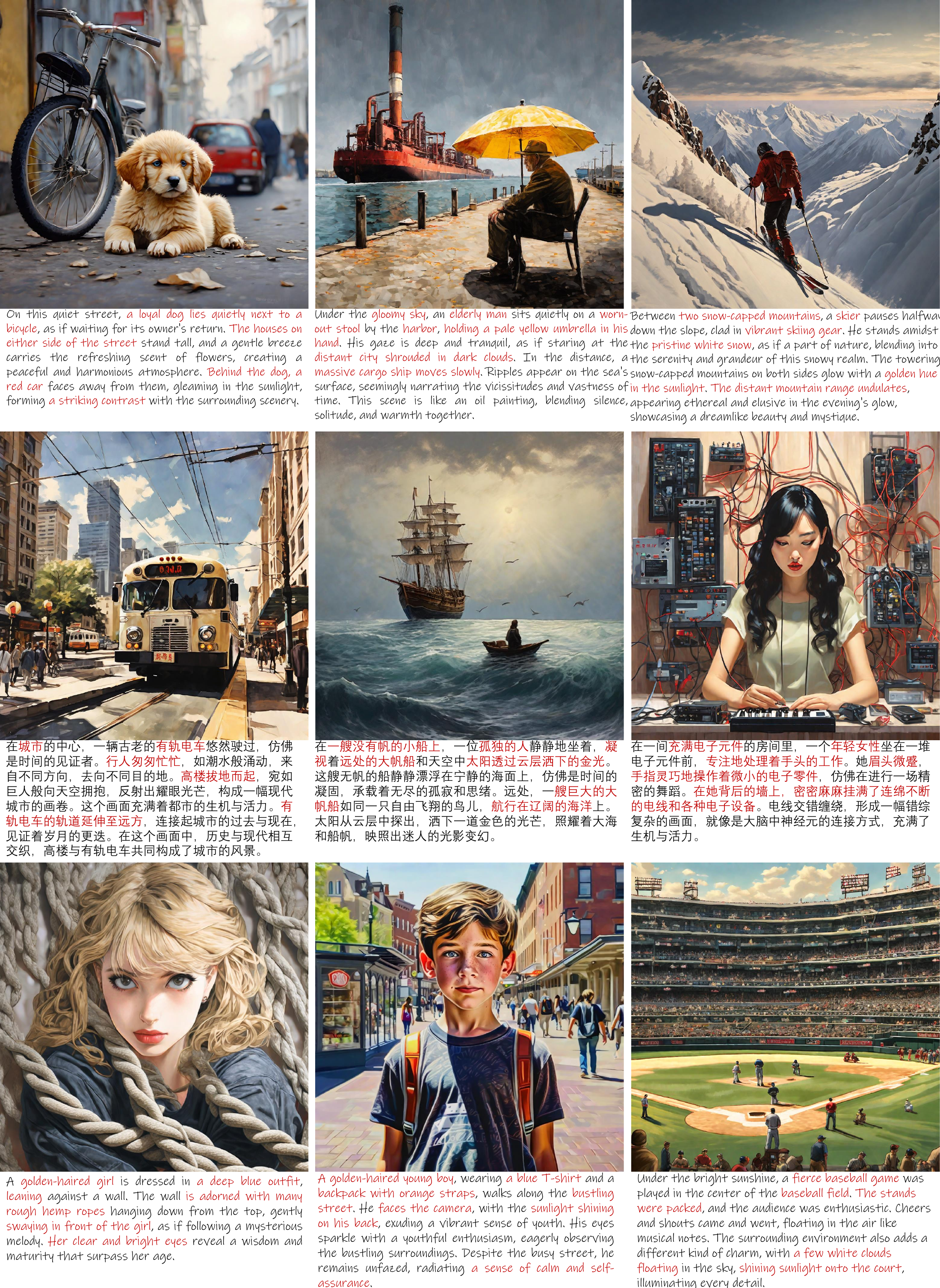}}
    \vspace{-3mm}
    \caption{\small Long prompts driven generation results.
    More examples showcasing that our model could capture long contextual text semantics and synthesize text-alignment results.
    Zoom in for details.
    }
    \label{fig:longprompts_sp_2}
    \vspace{-12pt}
\end{figure}

\begin{figure}[h]
    \centerline{\includegraphics[width=.89\linewidth]{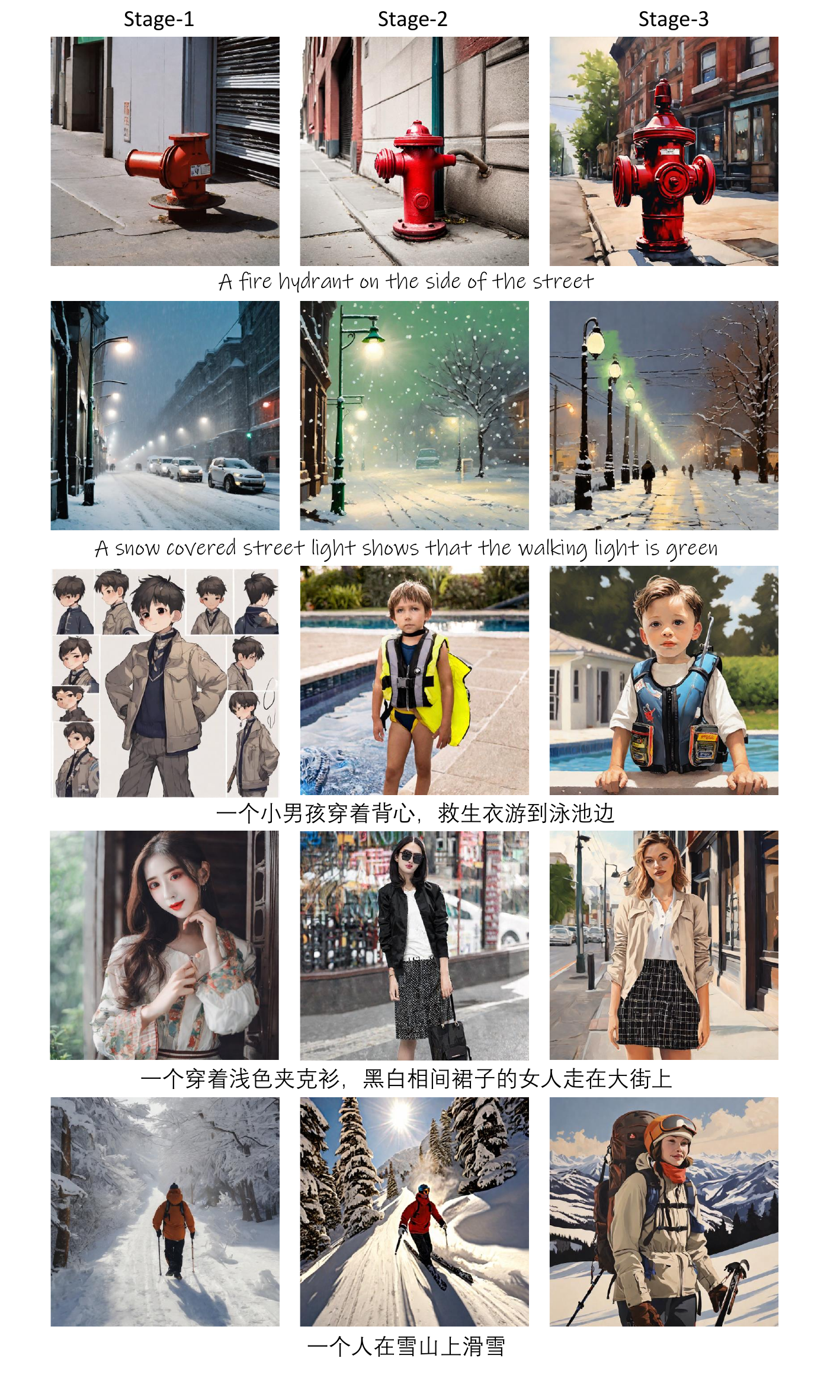}}
    \vspace{-4mm}
    \caption{\small Comparison results between stages.
    ``Stage i'' indicates the images generated by models trained after Stage i, where i$\in$[1, 2, 3].
    Clearly, the synthesis quality, including the overall image structure, fine-details, visual aesthetics improves with the training stages progresses.
    }
    \label{fig:stage_compare}
    \vspace{-12pt}
\end{figure}

\begin{figure}
    \centerline{\includegraphics[width=1\linewidth]{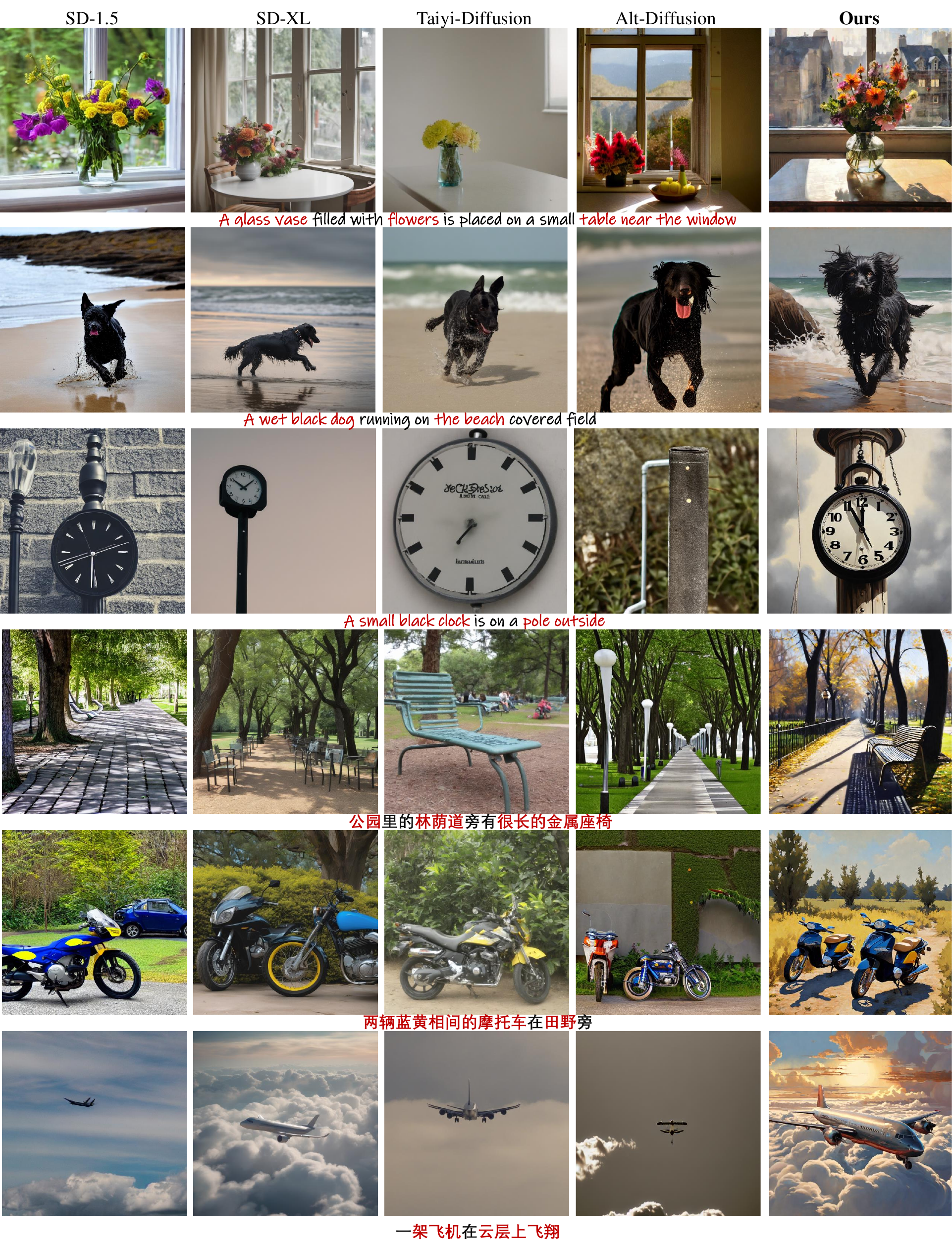}}
    \caption{\small More visual comparison results between our proposed method and popular text-to-image diffusion baselines.
    For models that do not support multilingual text conditions, we translate given prompts into corresponding language to generate images for comparison.
    Our proposed method could produce images that are plausible, with better text-image alignment and higher visual quality. 
    }
    \label{fig:sota_compare_showcase_sp}
    \vspace{-3mm}
\end{figure}

\begin{figure}
    \centerline{\includegraphics[width=1\linewidth]{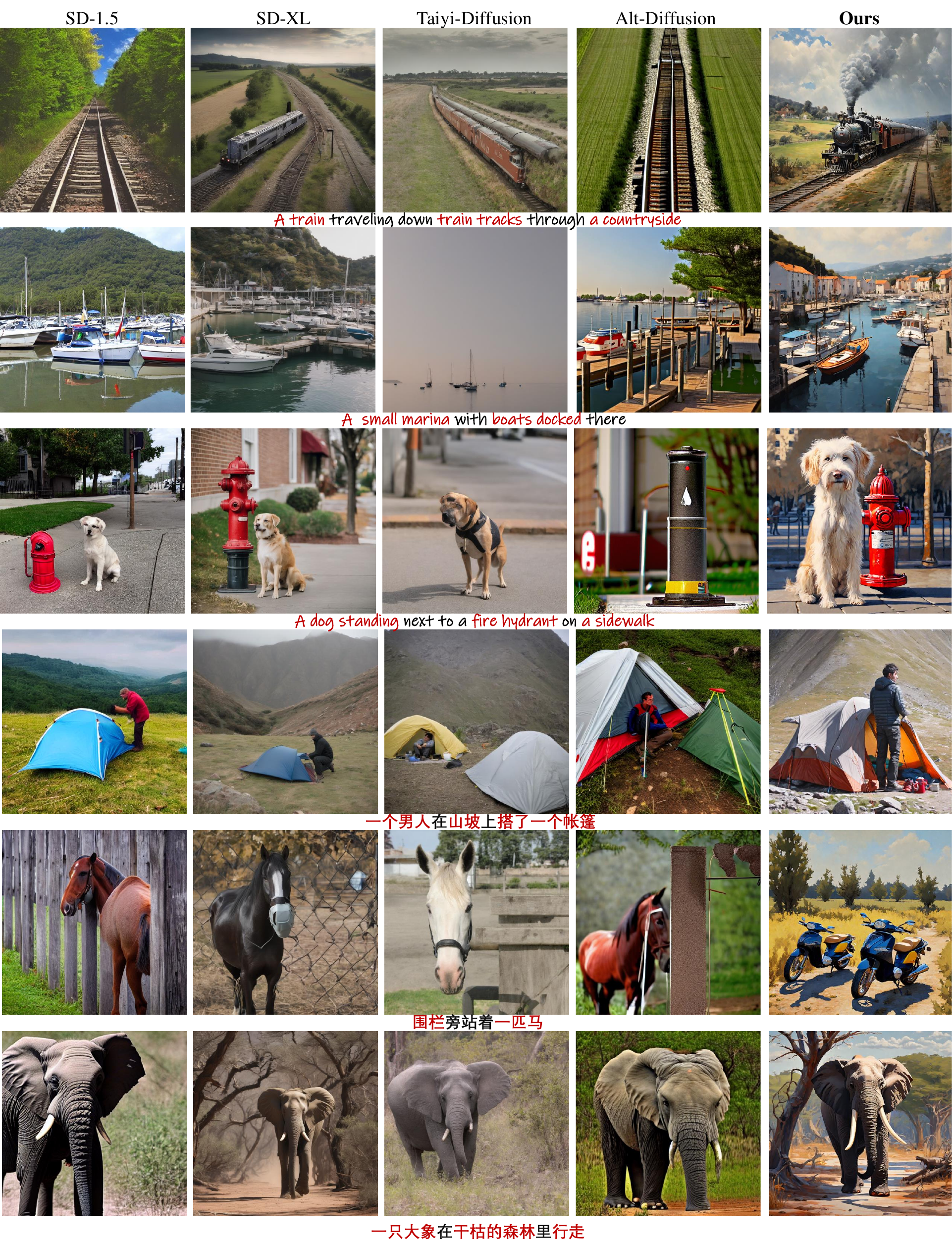}}
    \caption{\small More visual comparison results between our proposed method and popular text-to-image diffusion baselines.
    For models that do not support multilingual text conditions, we translate given prompts into corresponding language to generate images for comparison.
    Our proposed method could produce images that are plausible, with better text-image alignment and higher visual quality. 
    }
    \label{fig:sota_compare_showcase_sp_2}
    \vspace{-3mm}
\end{figure}

\end{document}